\documentclass[10pt, a4paper]{article}

\usepackage{hyperref}
\usepackage{url}

\usepackage{amsmath}
\usepackage{amscd}
\usepackage{amsthm}
\usepackage{amssymb}
\usepackage{amsfonts}
\usepackage{amsrefs}
\usepackage{enumitem} 
\usepackage{bm}
\usepackage{mathtools}
\usepackage{mathdots}
\usepackage[T1]{fontenc}
\usepackage{framed, color}
\usepackage{dsfont}
\usepackage{graphicx}
\usepackage{hyperref}
\usepackage{latexsym}
\usepackage{mathrsfs}
\usepackage{subfig}
\usepackage{verbatim}
\usepackage{booktabs} 
\usepackage{placeins}
\usepackage{thmtools, thm-restate}
\usepackage{multirow}
\usepackage{todonotes}

\theoremstyle{plain}
\newtheorem{theorem}{Theorem}[section]

\newtheorem{lemma}[theorem]{Lemma}

\newtheorem{remark}[theorem]{Remark}

\theoremstyle{definition}
\newtheorem{definition}[theorem]{Definition}

\newcommand{\RR}{\mathbb R}
\newcommand{\ZZ}{\mathbb Z}
\newcommand{\CC}{\mathbb C}
\newcommand{\NN}{\mathbb N}

\newcommand{\PP}{\mathbb{P}}

\newcommand{\XX}{\mathcal{X}}
\newcommand{\Ltwo}{L^{2}(\RR)}
\newcommand{\elltwo}{\ell^{2}(\ZZ)}

\newcommand{\dt}{\ \mbox{d}t}
\newcommand{\DD}{\mathcal{D}}

\title{Subpixel object segmentation using wavelets and multi resolution analysis}

\author{
	Ray Sheombarsing
	\\ \small{ {r.sheombarsing@nki.nl}}
	 \and   
	Nikita Moriakov
	\\ \small{  {n.moriakov@nki.nl}}
	\and   
	Jan-Jakob Sonke	 
	\\ \small{ {j.sonke@nki.nl}}
	 \and
	 Jonas Teuwen	 
	 \\ \small{ {j.teuwen@nki.nl}}
	 \\[2ex]
	Netherlands Cancer Institute \thanks{The Netherlands Cancer Institute, Amsterdam, The Netherlands, Plesmanlaan 121, 1066 CX Amsterdam.}
}

\begin{document}

\maketitle

\begin{abstract}
We propose a novel deep learning framework for fast prediction of boundaries of 
two-dimensional simply connected domains using wavelets and Multi Resolution Analysis 
(MRA). The boundaries are modelled as (piecewise) smooth closed curves using wavelets 
and the so-called Pyramid Algorithm. Our network 
architecture is a hybrid analog of the U-Net, where the down-sampling path is a 
two-dimensional encoder with learnable filters, and the upsampling
path is a one-dimensional decoder, which builds curves up from low to high resolution 
levels. Any wavelet basis induced by a MRA can be used. This flexibility allows 
for incorporation of priors on the smoothness of curves. The 
effectiveness of the proposed method is demonstrated by delineating boundaries of 
simply connected domains (organs) in medical images using Debauches wavelets and
comparing performance with a U-Net baseline. Our model demonstrates up to 5x faster inference speed 
compared to the U-Net, 
while maintaining similar performance in terms of Dice score and Hausdorff distance.	
\end{abstract}

\section{Introduction}
Semantic image segmentation is a core component of many medical imaging related tasks. 
Both as part of a pipeline to find a region of interest, or a task by itself, e.g., for measuring 
tumor volume. Nowadays, almost all segmentation algorithms in medical imaging are replaced 
by U-Net-like architectures \cite{unet} combining an encoder and decoder. Typically, the decoder 
is an upsampling path, and additional skip connections between the encoding and decoding part 
are added to recover the image's spatial information. While many variants or more exotics methods,
such as multi-scale and pyramid based approaches, recurrent networks or 
generative techniques, can be designed, all of these still yield per-pixel classifications. In these settings 
an image is interpreted as a \emph{discrete} collection of ordered pixels (or voxels in the three-dimensional case), 
where the task is to assign an appropriate class to each pixel using a probabilistic model. Typically, 
the pixels are assumed to be independent so that the joint-likelihood is tractable.

Medical images often represent discrete approximations, measured and stored in a discrete grid, of inherently 
continuous objects. The pixel or voxel sizes can be coarse, and each voxel may represent multiple tissue types 
due to the partial volume effect. While humans are able to draw continuous contours based on such images, e.g., 
boundaries of organs, typical deep learning-based segmentation models are not able to take this continuous 
nature into account. In this work, we propose a novel deep learning framework which does not predict per-pixel segmentation maps, 
but directly predicts contours as continuous objects. 

A fundamental component in the setup of our framework is the decision on how to
represent (closed) curves. While the Fourier basis is a natural candidate at first glance, its global nature may hamper
accurate predictions of curves which exhibit highly localized behavior, requiring accurate estimates of small noisy
high-frequency modes. For this reason, we have chosen to represent contours using wavelets and Multi Resolution Analysis
(MRA) instead. The main idea is to choose a single map $\varphi$, the so-called scaling function or father wavelet, and to
construct subspaces of functions associated to prescribed resolution levels by taking the span of appropriate dilations
and translations of $\varphi$. This setup 
provides an efficient way to decompose and reconstruct contours, from low to high resolution level, using
the classical Pyramid Algorithm \cite{Mallat}.

In this paper, we construct 
a hybrid analog of the U-Net, where the down-sampling path is a two-dimensional encoder with learnable filters, and the upsampling
path is an one-dimensional decoder, which builds curves up from low to high resolution levels. The filters in the decoder 
are not learned but uniquely determined by the chosen wavelet basis. Any wavelet basis induced by a MRA can be used. This flexibility allows 
for incorporation of priors on the smoothness of curves. 

\paragraph{Related work}
Previous work \cite{Chen2019,Marcos2018, Hatamizadeh2019} also proposed models  to predict contours by combining 
Active Contour Models (ACM) with a CNN into an end-to-end model. In these papers the representation of curves is
based on traditional pixel-based methodology. For instance, in \cite{Hatamizadeh2019} curves are modelled as level sets of 
distance maps defined on a discretization of the domain of the image. In \cite{Chen2019} a similar approach is followed, but a 
smoothed approximation of an indicator function is used instead of a distance map. The work in \cite{Marcos2018}
is perhaps most closely related to ours; they directly construct polygonal approximations of curves and represent them using (pixel) coordinates of the nodes.
The objectives minimized in the cited papers are based on a careful consideration of mean pixel intensities and geometric properties 
such as area and arc length. These properties are implicitly encoded in an objective function (energy-functional) defined on a space of 
distance maps \cite{Hatamizadeh2019}, suitable approximations of indicator functions \cite{Chen2019}, or family of polygons \cite{Marcos2018}. 
However, in contrast to our approach, the above methods all still provide pixel-based output. To the best of our knowledge, our work is the first 
to use MRA and wavelet analysis to construct pixel-independent representations of (closed) curves.

~\newline
This paper is organized as follows. 
In Section \ref{sec:setup} we review the mathematical background needed to construct our model. We explain 
how contours can be decomposed and reconstructed on different resolution levels. The reconstruction
algorithm, known as the Pyramid Algorithm, forms a core component of our network
architecture. In Section \ref{sec:model} we set up a model architecture and 
loss. Subsequently the datasets, training method and performance measures are 
described in Section \ref{sec:training}. We end the paper with the results of our method
in Section \ref{sec:results} and a brief discussion. A detailed account of all the mathematical
details is provided in the appendix. The code is available at \url{https://github.com/NKI-AI/mra_segmentation}.

\section{Background and mathematical setup}
\label{sec:setup}
Let $\left( \Omega, \Sigma, \PP \right)$ be a probability space and $X : \Omega \rightarrow \left[0, 1 \right]^{n_{1} \times n_{2}}$
a random variable whose realizations correspond to two dimensional gray-valued images, e.g., slices of MRI scans of size 
$n_{1} \times n_{2}$, where $n_{1}, n_{2} \in \NN$ and $\XX := \left[0, 1 \right]^{n_{1} \times n_{2}}$. We assume without loss of 
generality that the gray-values of the scans are rescaled to $[0,1]$. Commonly, $X$ is paired with a random variable 
$Y : \Omega \rightarrow \{0, 1\}^{n_{1} \times n_{2}}$ whose realizations correspond to binary segmentations. 
In this context, realizations of $(X,Y)$ correspond to scans with inscribed masks. In this paper we assume that the regions 
approximated by the inscribed masks are almost surely enclosed by \emph{simple closed curves}. 

Let $(x,y) \in \XX \times \{0, 1\}^{n_{1} \times n_{2}}$ be a realization of $(X,Y)$. The binary mask $y$ constitutes a discretized 
approximation of the true region of interest $R \subset \RR^{2}$ and provides a
polygonal approximation of $\partial R$. In a large variety of situation, however, $\partial R$ is in reality a closed (piecewise) smooth 
curve. 
In this paper we develop a deep learning framework for directly parameterizing $\partial R = \partial R(x)$, 
given an image $x$, using wavelets and Multi Resolution Analysis (MRA). Since wavelets play a seminal role 
in our set up, we start in the next section with an extensive review to introduce notation and terminology. 

\subsection{Multi resolution analysis}
\label{subsec: wavelets}
In this section we review the necessary material from discrete wavelet theory using the framework of a multi resolution analysis (MRA). 
We closely follow the exposition in \cite{HarmonicAnalysis} and \cite{WaveletsTheory}. The reader is referred to these references for a more comprehensive introduction.

\paragraph{Multi Resolution Analysis}
It is a well known-fact from Fourier analysis that a signal $\gamma \in \Ltwo$ cannot be simultaneously localized
in the time and frequency domain, see \cite{HarmonicAnalysis}. Multi resolution analysis aims to address this shortcoming by decomposing
a signal on different \emph{discrete} resolution levels. The idea is to construct subspaces $V_{j} \subset \Ltwo$, 
associated to various resolution levels $j \in \ZZ$, spanned by shifts of a localized mapping $\varphi_{j}$. 
The level of localization associated to $V_{j}$ is determined by taking an appropriate dilation of a prescribed map $\varphi$; the so-called
\emph{scaling function}. In the MRA framework the dilation factors are chosen to be powers of two.
Altogether, this yields an increasing sequence of closed subspaces $V_{j} \subset \Ltwo$ 
comprising all of $\Ltwo$. In particular, $V_{j+1} \supset V_{j}$ is the next level up in resolution after $V_{j}$.  
The ``view'' of $\gamma$ on resolution level $j$ is then defined as the orthogonal projection of $\gamma$ onto $V_{j}$. 
The formal definition of a MRA is given below.

\begin{definition}[Formal definition MRA]
\label{def:MRA}
Let $T_{k}: \Ltwo \rightarrow \Ltwo$ and $\DD_{j}: \Ltwo \rightarrow \Ltwo$ 
denote the translation and normalized dilation operator, respectively, defined by 
$T_{k}\gamma (x) = \gamma(x - k)$ and $\DD_{j} \gamma (t) = 2^{\frac{j}{2}} \gamma( 2^{j} t )$
for $\gamma \in \Ltwo \cap C^{\infty}_{0}(\RR)$ and $j, k \in \ZZ$. 
A multi-resolution analysis of $\Ltwo$ is an increasing subsequence of subspaces $(V_{j})_{j \in \ZZ}$, such that 
	$(i)$ $\bigcap_{j \in \ZZ} V_{j} = \{0\}$,
	$(ii)$ $\bigcup_{j \in \ZZ} V_{j}$ is dense in $\Ltwo$,
	$(iii)$ $\gamma \in V_{j}$ if and only if $\DD_{1} \gamma \in V_{j+1}$,
	$(iv)$ $V_{0}$ is invariant under translations,
	$(v)$ $\exists \varphi \in \Ltwo$ such that $\{ T_{k}\varphi \}_{k \in \ZZ}$
	         is an orthonormal basis for $V_{0}$. 
\end{definition}
Conditions $(iii)$, $(iv)$ and $(v)$ imply that each subspace $V_{j}$ is closed and invariant under integer shifts. 
Condition $(v)$ formalizes the idea that the subspaces are spanned by translations and dilations of a single map 
$\varphi$; the so-called \emph{scaling function} or \emph{father wavelet}. Indeed, it is straightforward to
show that $\{ \varphi_{jk} : k \in \ZZ \}$ is an orthonormal basis for $V_{j}$, where $\varphi_{jk} := T_{k}\DD_{j} \varphi$.

\paragraph{Approximation coefficients}
Let $\gamma_{j} \in V_{j}$ denote the best approximation of $\gamma$ at resolution level $j$, i.e., 
$\gamma_{j} := P_{j} \gamma$, where $P_{j} : \Ltwo \rightarrow V_{j}$ is the orthogonal projection onto $V_{j}$.
Then there exists unique coefficients $a_{j}(\gamma) = (a_{jk}(\gamma) )_{k \in \ZZ}$ such that 
$
	\gamma_{j} = \sum_{k \in \ZZ} a_{jk}(\gamma) \varphi_{jk},
$
where $a_{jk} = \langle \gamma, \varphi_{jk} \rangle$, 
since $\{ \varphi_{jk} : k \in \ZZ \}$ is an orthonormal basis for $V_{j}$. The coefficients $a_{j}(\gamma)$ 
are referred to as the \emph{approximation coefficients} of $\gamma$ at resolution level $j$. 
The associated subspaces $V_{j}$ are referred to as the approximation subspaces. We will write
$a_{j}(\gamma) = a_{j}$ whenever it is clear from the context that the coefficients are associated to $\gamma$. 

\paragraph{Detail coefficients}
To study the information that is lost when a signal in $V_{j+1}$ is projected onto $V_{j}$, we consider the operator
$Q_{j}: = P_{j+1} - P_{j}$. Note that $Q_{j} \gamma \in W_{j} := V_{j}^{\perp} \cap V_{j+1}$, since $P_{j} P_{j+1} = P_{j}$ 
and $V_{j} \subset V_{j+1}$. 
The subspace $W_{j}$ is called the detail subspace at level $j$. The detail subspaces 
$( W_{j})_{j \in \ZZ}$ are mutually disjoint and orthogonal by construction. Furthermore, since $V_{j} = V_{j-1} \oplus W_{j-1}$
for any $j \in \ZZ$, it follows that 
$V_{j} = V_{j_{0}} \oplus \bigoplus_{l=j_{0}}^{j-1} W_{l}$,
for all $j > j_{0}$, where $j, j_{0} \in \ZZ$. This decomposition shows that a signal on resolution level $j$ can be reconstructed
from any lower level $j_{0}$ if all the details in between are known and that 
$\Ltwo = \bigoplus_{j \in \ZZ} W_{j}$. 

A fundamental result, known as Mallat's Theorem, states that the subspaces $W_{j}$ can too be spanned by dilating
and shifting a single map. More precisely, there exists a map $\psi \in W_{0}$, the so-called \emph{mother wavelet}, such that
$\{ \psi_{jk} : k \in \ZZ \}$ is an orthonormal basis for $W_{j}$, where $\psi_{jk} := T_{k}\DD_{j} \psi$. The wavelet constructed
in Mallat's Theorem is often referred to as the \emph{mother wavelet}. 
The proof is based on a very careful analysis of the defining properties of a scaling function in the frequency domain.
The reader is referred to \cite{HarmonicAnalysis} for the details. Now, using Mallat's mother wavelet, we may write 
$Q_{j} \gamma = \sum_{k \in \ZZ} d_{jk}(\gamma) \psi_{jk}$ for any $\gamma \in \Ltwo$, where 
$d_{j}(\gamma) := (d_{jk}(\gamma))_{k \in \ZZ} \in \elltwo$. The coefficients $d_{j}(\gamma)$ are
referred to as the \emph{detail coefficients} of $\gamma$ at resolution level $j$. We will suppress the dependence on 
$\gamma$ and write $d_{j}(\gamma) = d_{j}$ when there is no chance of confusion. 

The detail coefficients store the information needed to go back one level up in resolution, since 
$P_{j+1} = P_{j}  + Q_{j}$ by construction. In general, given approximation
coefficients $a_{j_{0}} \in \ell^{2}(\ZZ)$ at level $j_{0}$ and detail coefficients $d_{l} \in \ell^{2}(\ZZ)$ at
levels $j_{0} \leq l \leq j-1$, we can reconstruct the approximation at level $j$ from $j_{0}$ via 
$
	\gamma_{j} = \sum_{k \in \ZZ} a_{j_{0}k} \varphi_{j_{0}k} + \sum_{j_{0} \leq l \leq j - 1} \sum_{k \in \ZZ} d_{lk} \psi_{lk}.
$

\paragraph{Galerkin projections}
In practice, we only approximate a finite number of approximation and detail coefficients. For this purpose, we 
introduce some additional notation. We define the finite dimensional
subspaces $\widehat{V}_{j} := \mbox{span} \{ \varphi_{jk} : - 2^{j-1} \leq k \leq 2^{j-1} -1 \} \subset V_{j}$ and
$\widehat{W}_{j} := \mbox{span} \{ \psi_{jk} : - 2^{j-1} \leq k \leq 2^{j-1} - 1\} \subset W_{j}$. We shall frequently 
identify $\widehat{V}_{j} \simeq \RR^{2^{j}}$ and $\widehat{W}_{j} \simeq \RR^{2^{j}}$.

\subsection{The Discrete Wavelet Transform}
In this section we recall how to compute the approximation and detail coefficients given
a prescribed scaling function $\varphi$. Many fundamental aspects of 
MRA's, both theoretical and computational, can be traced back to the following simple but key observation. 
Since $V_{0} \subset V_{1}$, there must exist coefficients $h \in \ell^{2}(\ZZ)$ such that 
$\varphi = \sum_{l \in \ZZ} h_{l} \varphi_{1l}$. This equation is referred to as the \emph{scaling equation};
one of the fundamental properties of a scaling function. The coefficients $h$ completely characterize the
scaling function. Therefore, to define a MRA, one only needs to specify an appropriate sequence $h \in \ell^{2}(\ZZ)$. 
Throughout this paper we assume that an appropriate choice for $h$ is given. 

\paragraph{Decomposition}
The scaling equation can be used to derive an efficient scheme for computing lower
order approximation coefficients (of any order) given an initial approximation $a_{j+1} \in \elltwo$. 
Define $C_{\tilde h}: \ell^{2}(\ZZ) \rightarrow \ell^{2}(\ZZ)$ and $M^{-}: \ell^{2}(\ZZ) \rightarrow \ell^{2}(\ZZ)$ 
by $C_{\tilde h}(a) = a \ast \tilde h$ and $(M^{-}a)_{k}= a_{2k}$, respectively, where $\tilde h \in \ell^{2}(\ZZ)$ is defined by 
$\tilde h_{k} = \overline{h_{-k}}$, and $\ast : \ell^{2}(\ZZ) \times \ell^{2}(\ZZ)  \rightarrow \ell^{2}(\ZZ)$ is
the two-sided discrete convolution. Then $a_{j} = M^{-}C_{\tilde h} a_{j+1}$ for any $j \in \ZZ$. 
Furthermore, one can show, by analyzing the Fourier transform of the scaling equation, that $C_{\tilde h}$ cuts off 
relatively high frequencies. For this reason, $h$ is often referred to as the \emph{low-pass filter} associated to the MRA, 
see \cite{HarmonicAnalysis, WaveletsTheory}

Similarly, since $\psi \in W_{0} \subset V_{1}$, there exist coefficients $g \in \ell^{2}(\ZZ)$ such that 
$\psi = \sum_{k \in \ZZ} g_{k} \varphi_{1k}$. For Mallat's mother wavelet, we have $g_{k} = (-1)^{k-1} \overline{h_{1-k}}$. 
An analogous argument shows that $d_{j} = M^{-}C_{\tilde g} a_{j+1}$ for all $j \in \ZZ$. Moreover, it can be shown that $C_{\tilde g}$ cuts of relatively
low frequencies, and is therefore referred to as the \emph{high pass filter} associated to the MRA. 

\paragraph{Reconstruction}
The orthogonal decomposition $V_{j+1} = V_{j} \oplus W_{j}$ can be used to reconstruct
$a_{j+1} \in V_{j+1}$ given the approximation and detail coefficients $a_{j} \in V_{j}$ and $d_{j} \in W_{j}$
at resolution level $j$. More precisely, substitution of the scaling equations for $\varphi$ and $\psi$ into this decomposition
yields $a_{j+1} =  C_{h} M^{+}a_{j} + C_{g} M^{+}d_{j}$, 
where $M^{+}: \elltwo \rightarrow \elltwo$ is a right inverse of $M^{-}$ defined by $(M^{+}a)_{k} := a_{\frac{k}{2}}$
if $k$ is even and zero otherwise. 

~\newline
The reconstruction and decomposition formulae together form the well-known Pyramid Algorithm \cite{Mallat}. 
In practice, our signals are periodic and do not directly fit into the MRA framework, since 
non-zero periodic signals are not elements in $\Ltwo$. We will address this issue in the next section
by using an appropriate cut-off. Here we only remark that the periodicity needs to be carefully taken into 
account in the decomposition and reconstruction formulae, see Appendix \ref{appendix:implementation}. 

\subsection{Wavelet representation of periodic curves}
\label{sec:init_approx_coeffs}
Let $j_{1}\in \NN$ be a prescribed initial resolution-level on which a periodic signal $\gamma$ is to be approximated. 
We re-parameterize $\gamma$ to have period $1$, so that we can conveniently choose the number 
of approximation coefficients to be a power of two (see the discussion in Appendix \ref{appendix: init_approx_coeffs}). 
Furthermore, to address the issue that periodic signals are not contained in $L^{2}(\RR)$, we restrict the 
re-parameterized curve to $[-1, 1]$, i.e., set $\gamma^{\ast}(t) := \gamma(lt) \bold{1}_{[-1, 1]}(t)$. In general, 
this will introduce discontinuities at the boundary points $-l$ and $l$. However, this issue is
easily dealt with, as described below, since we only need information about $\gamma$ on a strict subset 
$[I_{0}, I_{1}] \subset [-l, l]$ of length $l$.

If $j_{1}$ is sufficiently large, the approximations
coefficients needed to (approximately) cover $[-1, 1]$ are $(a_{j_{1}k}(\gamma^{\ast}))_{k= - 2^{j_{1}}}^{\lfloor 2^{j_{1} - \beta} \rfloor}$.
Here $\beta > 0$ is the support of the underlying wavelet. 
These coefficients will be close to the (scaled) sample values of $\gamma$ on 
$\{ k 2^{-j_{1}}: - 2^{j_{1}} \leq k \leq \lfloor 2^{j_{1}} - \beta \rfloor \}$. We refer to 
Appendix \ref{appendix: init_approx_coeffs} for an explanation. 
Motivated by this observation, we
use the coefficients $(a_{j_{1}k}(\gamma^{\ast}))_{k= -2^{j_{1} -1 }}^{2^{j_{1} - 1} - 1}$ only, which cover
$[-\frac{1}{2}, \frac{1 - 2^{1 - j_{1}}}{2}]$ approximately.
To ensure that $2^{j_{1} -1} -1 < \lfloor 2^{j_{1}} - \beta \rfloor$, we require that
$j_{1} \geq \left \lceil \frac{ \log{(\beta - 1)} }{ \log(2) } + 1 \right \rceil$. 
This quantity is well-defined for all bases considered in this paper, except the Haar-basis.

Formally, the approximation coefficients at a prescribed level $j$ need to be computed by evaluating
the integrals $\langle \gamma^{\ast}, \varphi_{jk} \rangle$. Typically, in signal analysis one uses appropriately 
rescaled samples of $\gamma^{\ast}$ instead. This approach is justified for signals sampled on a sufficiently
high resolution, since the family $(\varphi_{j0})_{j \in \ZZ}$ behaves as an approximate identity as 
$j \rightarrow \infty$. Such an initialization scheme may be highly inaccurate, however, when the signal
is not sampled on a sufficiently fine grid. In the appendix, we propose an alternative to the usual initialization 
scheme by exploiting the fact that we are dealing with periodic signals; see Lemma \ref{lemma:approx_coeffs}. 

\section{Model}
\label{sec:model}
In this section we formulate the objective of our model and present a suitable architecture. Furthermore, 
we set up a loss function for determining network-parameters. We will slightly 
abuse notation from the previous sections. A contour associated to an image $x \in \XX$ will be denoted by $\gamma(x)$
and is henceforth assumed to have two components denoted by $[\gamma(x)]_{1}$ and $[\gamma(x)]_{2}$. 
In general, for any quantity that depends on $x$, we will frequently denote its dependency on $x$ in a similar way. 
Finally, all operations from the previous sections are understood to be carried out component-wise. 

\subsection{Objective}
Let $y \in \mathcal{Y}$ be a mask associated to an image $x \in \XX$, where we regard $y$ as a realization of 
$Y \vert X = x$. We assume that the boundary $\partial R(x)$ of the region $R(x)$ encoded in $y$ is parameterized 
by a closed curve $\gamma(x) \in C^{2}_{\text{per}}([0, l(x)]; \RR^{2})$. Here $l(x)>0$ denotes the arc-length of $\gamma(x)$. 
The objective is to compute the approximation coefficients of $\gamma^{\ast}(x)$, where $\gamma^{\ast}(x)$ denotes the 
truncated re-parameterization of $\gamma(x)$. More precisely, let $j_{0}, j_{1}, j_{2} \in \NN$ be resolution levels,
where $j_{0} \leq j_{1} \leq j_{2}$. We will construct a convolutional neural network 
$F: \XX \times \Theta \rightarrow (0, 1) \times \prod_{j=j_{0}}^{j_{2}} \widehat{V}_{j} \times \widehat{V}_{j}$, where
$F(x, \theta) = \left( p(x, \theta), f_{j_{0}}( x, \theta), \ldots, f_{j_{2}}(x, \theta) \right)$. Here $p(x, \theta)$ is the
conditional probability that $R(x) \not = \emptyset$, $f_{j}(x, \theta)$ is a prediction for the approximation coefficients
of $\gamma^{\ast}(x)$ at level $j$, and $\theta \in \Theta \subset \RR^{n}$ are the network parameters. 
The goal is to find $\theta \in \Theta$ such that $f_{jk}(x, \theta) \approx \left( a_{jk} ( [\gamma^{\ast}(x)]_{1}), a_{jk} ( [\gamma^{\ast}(x)]_{2}) \right)$ 
for $-2^{j -1} \leq k \leq 2^{j - 1} - 1$, $j_{0} \leq j \leq j_{2}$ and ``most'' realizations of $X$. 
Henceforth we will omit the dependence of $F$ and its components on $\theta$.  

\subsection{Network architecture}
Our network architecture consists of an encoder, bottleneck and decoder 
with skip-connections in between. Only the approximation coefficients at the lowest resolution-level 
$j_{0}$ are ``directly'' computed by the network (in the bottleneck). Afterwards, the Pyramid Algorithm takes over
to compute approximation coefficients on higher resolution levels (the decoder). In practice, the detail coefficients are 
negligible on sufficiently high resolution levels. For this reason, we only predict detail coefficients up to a 
prescribed level $j_{1}$. The predictions at higher resolution levels $j_{1} < j \leq j_{2}$ are computed without detail 
coefficients. 

The encoder consists of five down-sampling blocks. Each block consists of six (convolutional) residual blocks, 
using ReLU-activation and kernels of size $3 \times 3$, followed by an average-pooling layer of size $2 \times 2$. 
The number of filters used in the five blocks, from top to bottom, is $32$, $32$, $64$, $64$, $128$, respectively. 

The encoder is followed by a a bottleneck. The first layer in the bottleneck is fully connected and transforms the output of the encoder to 
a vector in $\RR^{128}$. Attached to this layer are five branches to predict the probability $p(x) \in (0, 1)$,
components $[f_{j_{0}}(x)]_{s} \in \widehat{V}_{j_{0}}$ and $[h_{j_{0}}(x)]_{s} \in \widehat{W}_{j_{0}}$,
where  $s \in \{1, 2\}$. The latter quantity $[h_{j_{0}}(x)]_{s}$ is a prediction for the detail coefficients 
at level $j_{0}$. Each branch consists of four fully-connected layers. The first three layers map 
from $\RR^{128}$ to itself with ReLu-activation and residual connections in between. The final fourth layer 
transforms the $128$-dimensional output to an element in $\RR^{n}$, where $n = 2^{j_{0}}$ 
for the branches associated to the approximation and detail coefficients, and $n=1$ for the probability.
The branch associated to the probability is followed by a sigmoid-activation. 

The detail coefficients at levels $j_{0} \leq j < j_{1}$ are predicted using skip-connections. More precisely,
the feature map from the encoder path, that corresponds to the same ``spatial level'', is compressed to a 
feature map with four channels using a $1 \times 1$ convolution. Subsequently, a branch of the above form
is used to predict detail coefficients in $\RR^{2^{j}} \simeq \widehat{W}_{j}$. Altogether, the
predicted approximation coefficients at level $j_{0}$ and detail coefficients at levels $j_{0} \leq j \leq j_{1} -1$ 
are used as input to the Pyramid algorithm and together form the upsampling path. 

\subsection{Loss}
The loss $\mathcal{L}$ consists of two parts: ordinary cross-entropy for optimizing the likelihood of $p(x)$
and a part corresponding to the $L^{2}$-error between observed and predicted curves 
on different resolution levels. More precisely, set $r(x) = 1$ if $R(x) \not = \emptyset$ and $r(x) = 0$ otherwise,
then $\mathcal{L}_{\text{ce}}( p(x), r(x) ) := r(x) \log p(x) + (1- r(x)) \log( 1 - p(x))$.
Next, suppose $a( [\gamma^{\ast}(x)]) = \left(a_{j_{0}}([\gamma^{\ast}(x)]), \ldots, a_{j_{2}}( [\gamma^{\ast}(x)])  \right)
\in \prod_{j=j_{0}}^{j_{2}} \widehat{V}_{j} \times \widehat{V}_{j}$ are the approximation coefficients of $\gamma^{\ast}(x)$
on resolution levels $j_{0} \leq j \leq j_{2}$. Define
$
	\mathcal{L}_{js}(f_{j}(x), a_{j}( \gamma^{\ast}(x))) :=
	\Vert [f_{j}(x)]_{s} -  a_{j}( [\gamma^{\ast}(x)]_{s}) \Vert_{2}
$
for $s \in \{1, 2\}$. This quantity corresponds to the $L^{2}$-error on resolution level $j$ between the curves 
with approximation coefficients $[f_{j}(x)]_{s}$ and $a_{j}( [\gamma^{\ast}(x)]_{s})$. Finally, define the total loss by 
\begin{align*}
	\mathcal{L}(F(x), a) := w \mathcal{L}_{\text{ce}}( p(x), r(x) ) + 
	\sum_{\substack{s \in \{1, 2\} \\ j \in \{j_{0}, j_{2} \}} }\mathcal{L}_{js}(f_{j}(x), a_{j}( \gamma^{\ast}(x))),
\end{align*}
where $w >0$ is a weight. In practice, we set $w = 1$.
Notice that $\mathcal{L}$ measures the discrepancies between observed and predicted curves on the lowest
and highest resolution levels only. 
In practice, this enforces the 
approximation and detail coefficients at intermediate levels to agree as well; see the experiments in Section \ref{sec:results}. 

\section{Training}
\label{sec:training}

In this section we describe the dataset on which we test our method. 
In addition, we provide details about preprocessing steps and model development (training). 

\subsection{Datasets}
\label{subsec:data}

\paragraph{Toy dataset}
The main purpose of the toy-example is to create a setting in which the annotated contours
differ substantially from annotations confined to a grid. For this purpose, we 
consider piecewise smooth curves having a finite number of non-differentiable points. 
The toy-dataset consists of hypocycloids, up to an Euclidian motion and scaling, defined by 
\begin{align*}
	\eta(t) := \begin{bmatrix} (r_{1} - r_{2}) \cos t + r_{2} \cos \left( \dfrac{r_{1} - r_{2}}{r_{2}} t \right) &
		(r_{1} - r_{2}) \sin t - r_{2} \sin \left( \dfrac{r_{1} - r_{2}}{r_{2}} t \right) \end{bmatrix}^{T}, 
\end{align*}
where $r_{1} > r_{2}$ and $t \in \RR$. If $\frac{r_{1}}{r_{2}} \in \NN$, 
then $\eta$ is closed and has exactly $\frac{r_{1}}{r_{2}}$ cusps (non-differentiable points). 
To easily control the number of cusps, we fix $r_{2} = 1$ and vary $r_{1}$. Note that in this case 
$\eta$ has $r_{1}$ cusps and period $2 \pi$. 

We construct curves and binary masks of various sizes, orientation and positions, by sampling
a radius $r_{1} \in \{3, 4, 5 ,6\}$ from $\mathcal{U}(\{3, 4, 5 ,6\})$, angle $\theta$ from $\mathcal{U}([- \frac{\pi }{2}, \frac{\pi }{2}])$, components
$q_{1}, q_{2}$ from $\mathcal{U}([-80, 80])$ for a shift, and a scaling factor $\kappa$ from $\mathcal{U}([10, 20])$. Here $\mathcal{U}(I)$ denotes
a random variable uniformly distributed on $I$, where $I$ is an interval of finite length or a discrete finite set. 
Next, we evaluate the curve $\kappa \left( R(\theta) \eta + \begin{bmatrix} 160 + q_{1} &  160 + q_{2} \end{bmatrix}^{T} \right)$ on an equispaced grid of $[0, 2 \pi]$ of
size $512$. Here $R(\theta)$ corresponds to an anti-clockwise rotation around the origin with angle $\theta$. Finally, the discretized
curve is used to construct a binary mask of size $320 \times 320$ using \textsc{skimage}. 

\paragraph{Medical decathlon}
The data used to evaluate the performance of our model consists of MRI images of the prostate and CT scans of the spleen. 
The datasets are part of a public dataset made available for the Medical Decathlon Contest  \cite{MedicalDecathlon} . The dataset for the prostate consists of
T$2$-weighted MRI images of size $320 \times 320$, which were cropped to size $224 \times 224$. The dataset for the spleen consists 
of CT scans of size $512 \times 512$ and was cropped to size $256 \times 256$. The cropping was based on constructing bounding
boxes of the form $[u_{\text{min}} - \delta_{p}, u_{\text{max}} +  \delta_{p}] \times [v_{\text{min}} - \delta_{p}, v_{\text{max}} +  \delta_{p}]$
for the training set, where $u_{\text{min}}, v_{\text{min}}, u_{\text{max}}$ and $v_{\text{max}}$ are the minimal and maximal coordinates of the 
segmentation in each direction, respectively, using an offset of $\delta_{p} = 65$ pixels.  
A residual CNN (encoder of five blocks) was trained (and validated) on the training set to regress the corner and center points of the 
bounding boxes using a RMSE-loss. This rather crude approach is not meant to produce tight bounding boxes, but serves as a
(rough) localization step to improve performance, and allows us to focus on the task of shape-prediction only. 

\subsection{Construction ground-truth}
In this section we describe how the ground-truth data is generated using the Pyramid Algorithm and Lemma \ref{lemma:approx_coeffs}. 
Let $(x, u) \in \XX \times \RR^{n_{s} \times n_{p}}$ be an image (slice) - contour pair, where $x$ is a slice of the CT or MRI scan, 
$u$ is a finite sequence of points approximating a closed curve and $n_{s}=2$ is the number of spatial components. 
Since we only have access to binary masks (for the public datasets), and not to the raw annotations themselves, we extract $u$ 
using \textsc{opencv}. While not ideal, we stress that $u$ contains ``subpixel'' information and is \emph{not} constrainted to an integer-valued grid.

\paragraph{Fourier coefficients}
To initialize the Pyramid Algorithm, we compute the approximation coefficients at level $j_{1}$ using Lemma \ref{lemma:approx_coeffs}.
To accomplish this, we need to compute a Fourier expansion for $u$. First, we parameterize the contour 
by arc length. The arc length $l$ is approximated by summing up the Euclidian distances between subsequent
points in $u$. The Fourier coefficients are then computed by evaluating the contour on an equispaced grid of $[0, l]$ of size $2N-1$, where $N \in \NN$, 
using linear interpolation and the Discrete Fourier Transform. Since the contours are real-valued, we only store the Fourier coefficients 
$(\tilde \gamma_{m})_{m=0}^{N-1} \in (\CC^{n_{s}})^{N}$. Fourier coefficients that are too small, i.e., have no relevant contribution, 
are set to zero; see Appendix \ref{appendix:truncation} for the details. 

\paragraph{Consistency}
To have consistent parameterizations for all slices, we ensure that $u$ is always traversed anti clock-wise (using \texttt{opencv}).
Furthermore, since the parameterization is only determined up to a translation in time, we need to pick out a specific one. 
We choose the unique parameterization such that the contour starts at angle zero at time zero relative to the midpoint 
$c = (c_{1}, c_{2} ) \in \RR^{2}$ of $R$. The implementation details are provided in Appendix \ref{appendix:midpoint}. 

\paragraph{Approximation and detail coefficients}
Altogether, the above steps yield a contour $\gamma$ with Fourier coefficients $(\gamma_{m})_{m=1-N}^{N-1}$ and period $l$. 
We reparameterize $\gamma$ to have period $1$, as explained in Section \ref{sec:init_approx_coeffs}, 
and ``center'' the contour using the average midpoint computed over the training-set.  
The initial approximation coefficients at level $j_{2}$ are then computed using Lemma \ref{lemma:approx_coeffs}. Next, the Pyramid Algorithm
is used to compute approximation and detail coefficients at levels $j_{0} \leq j \leq j_{2} - 1$. We set
the detail coefficients which are in absolute value below $\varepsilon = 5 \cdot 10^{-3}$ to zero to reduce noise.
Subsequently, we reconstruct the approximation coefficients
at levels $j_{0} + 1 \leq j \leq j_{1}$ using the thresholded detail coefficients. No detail coefficients are used to compute approximation coefficients
at levels $j_{1} + 1 \leq j \leq j_{2}$.
The final approximation and (thresholded) detail coefficients $a \in \bigoplus_{j=j_{0}}^{j_{2}}  \widehat{V}_{j} \times \widehat{V}_{j}$ and 
$d \in \bigoplus_{j=j_{0}}^{j_{1} - 1}  \widehat{W}_{j} \times \widehat{W}_{j}$ , respectively, are used as ground-truth. 

The resulting dataset $\mathcal{D}$ thus consists of tuples $( x, a, d)$.
The training and validation set $\mathcal{D}_{\text{train}}$ and $\mathcal{D}_{\text{val}}$, respectively, are obtained 
by randomly omitting subjects from the the full dataset $\mathcal{D}$. 
For the toy-example, spleen and prostate we have 
$(\vert \mathcal{D}_{\text{train}} \vert, \vert \mathcal{D}_{\text{val}} \vert) = (1650, 250)$, 
$(\vert \mathcal{D}_{\text{train}} \vert, \vert \mathcal{D}_{\text{val}} \vert) = (527, 75)$,
and
$(\vert \mathcal{D}_{\text{train}} \vert, \vert \mathcal{D}_{\text{val}} \vert) = (3148, 502)$, respectively. 
Before feeding the images $x$ into the model, we linearly rescale the
image intensities at each instance to $[0,1]$. Furthermore, for the spleen and prostate, we use extensive data augmentation: we use random shifts,
random rotations, random scaling, elastic deformations, horizontal shearing and random cropping. 

\subsection{Model training}
We use the Adam optimizer \cite{Adam} to train our network for $150$ epochs. 
In the first five epochs, we use a linearly increasing learning rate from $5 \cdot 10^{-4}$ to $10^{-3}$, 
which is subsequently decayed by a factor of $0.5$ each time the loss does not significantly decrease 
for $10$ subsequent epochs (for the remaining $145$ epochs).  A batch size of $16$ samples is used in each descent step. 
Finally, the model with the lowest loss is selected. The computations were performed in \textsc{PyTorch} on a Geforce RTX $2080$ Ti.

\begin{figure}[!thb]
	\centering
	\includegraphics[width=0.98\textwidth]{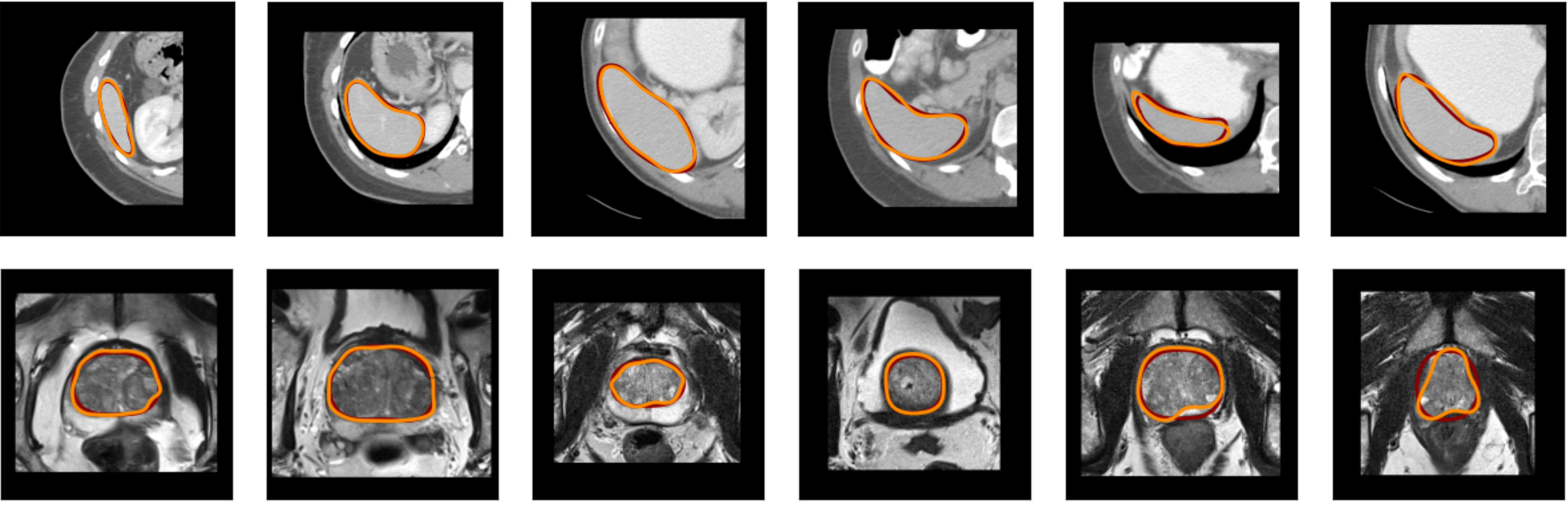}
	\caption{Predicted and observed boundaries colored in red and orange, respectively, for the spleen and prostate
		      for the best models (db$8$). The last two columns correspond to ``hard'' examples. Predictions for the 
		      other wavelet bases, as well as the toy-example, can be found in Appendix \ref{appendix:visualization}.}
	\label{fig:example_pred}
\end{figure}

\section{Results}
\label{sec:results}
In this section we examine the performance of our method. We investigate the dependence
on the choice of basis  and smoothness; we consider the Debauches wavelets db$p$ with $p \in \{1, 2, 4, 8, 16 \}$ vanishing 
moments. Roughly speaking, db$p$ corresponds to the unique MRA for which the mother wavelet has minimal support
and $p$ vanishing moments. For each basis, we fix appropriate resolution levels $j_{0}$, $j_{1}$ and $j_{2}$ as follows. 
We choose $j_{0}$ as the smallest level for which $2^{j_{0}-1}$ is 
larger or equal to the order of the low-pass-filter. Since the low-pass
filter associated to db$p$ has order $2p$, we set $j_{0}(p) = p$. 
We choose $j_{1} > j_{0}$ as the smallest level for which 
the detail coefficients are below $\varepsilon =5 \cdot 10^{-3}$ for all samples in the training set. 
We choose $j_{2} \geq j_{1}$ as the smallest resolution level for which the distance between the end-points
of the curves are within $1$ pixel distance for all samples in the training set.  Finally, 
we use $64$, $32$ and $64$ Fourier coefficients for the the toy-problem, spleen and prostate, respectively, 
to initialize the approximation coefficients. 

\paragraph{Baseline}
For the prostate and spleen we use the original two-dimensional U-Net in \cite{unet} as baseline with the following
differences: we use $32$ filters in the first layer instead of $64$ and use group normalization with four groups. 

\paragraph{Performance measures}
We evaluate accuracy using two-dimensional quantities only, since our models are $2$d. We compute the component-wise $L^{2}$-errors
between observed and predicted curves on the highest resolution level $j_{2}$ by 
taking the $\ell^{2}$-norm of the approximation coefficients. Furthermore, we compute the dice score and Haussdorf distance between
curves using the implementation in \textsc{shapely}. This requires a polygonal approximation of the contour, which is easily obtained 
using the approximation coefficients at level $j_{2}$.
Note that the Hausdorff distance between subsets of a general metric space (baseline) may differ from the Hausdorff 
distance between the associated boundaries (wavelet models), but coincide for (compact) simply connected subsets of $\RR^{n}$. 
The results are shown in Table \ref{table:results}. The predictions for the best performing wavelet models with respect to the 
Hausdorff-score are shown in Figure \ref{fig:example_pred}. Predictions for other bases, corresponding wavelet decompositions, 
and detailed visualization of statistics (violin plots) can be found in Appendix \ref{appendix:visualization}.

 \paragraph{Toy problem}
We observe that all models perform well and are capable of accurately parameterizing piecewise smooth curves.
For db$1$, however, we observe relatively large gaps between the end-points, since its support is relatively small.
In addition, db$1$ has difficulty with accurately predicting small ``densely'' sampled cycloids. 
In general, the predictions associated to the less regular wavelets db$1$ and db$2$ sometimes exhibit ``small'' oscillatory behavior.
We found that the latter two issues were caused by a too large resolution level $j_{2}$. 
To see why, note that the features extracted from the images only contain information up to a 
certain resolution level. The subpixel information needed to ``fill in the blanks'', so to speak, is in part provided by the ground-truth
data and in part by the chosen wavelet basis. The regularity and support of the wavelet determines
to which extent, i.e., up to which resolution level, subpixel information can be ``filled in''. 
As the regularity (and support) of the wavelet decreases, the maximal achievable resolution level decreases as well. 
\begin{table}[!tbp]
	\centering
		\captionof{table}{Mean and standard deviation of various performance measures for the toy example,
		spleen and prostate.  The standard deviation is reported in parentheses. The column
		$N_{p}$ is the approximate number of model parameters in millions, 
		$T$ is the inference time per image in milliseconds, and db$p$- refers to a network trained without detail coefficients.
		The length of the encoder for the toy-example is six and five for the prostate and spleen.} 
		\scalebox{0.87}{
		\begin{tabular}[htb]{ccccccccc}
		\toprule
		& Model & $(j_{0}, j_{1}, j_{2})$ & Dice & Hausdorff & $L^{2}$ ($s=1$) & $L^{2}$ ($s=2$) & $N_{p}$ & $T$ \\
		\midrule
		\multirow{6}{*}{\rotatebox{90}{Toy \hphantom{blablabla}}}
		& db$1$  & $(1, 6, 8)$ &  $0.962$ $(0.028)$  &  $3.520$ $(1.804)$  &  $0.846$ $(0.317)$  &  $0.830$ $(0.281)$  & $7.70$ & $15.6$ \\
		& db$2$  & $(2, 6, 8)$ &  $0.971$ $(0.021)$  &  $2.330$ $(0.889)$  &  $0.377$ $(0.172)$  &  $0.373$ $(0.161)$ & $5.96$ & $15.1$ \\
		& db$4$  & $(3, 6, 8)$ &  $\bf{0.980}$ $(0.014)$  &  $1.317$ $(0.509)$  &  $0.264$ $(0.122)$  &  $0.267$ $(0.124)$ & $5.45$ & $14.5$ \\
		& db$8$  & $(4, 6, 8)$ & $0.978$ $(0.016)$  &  $\bf{1.287}$ $(0.520)$  &  $0.276$ $(0.149)$  &  $0.281$ $(0.162)$ & $5.25$ & $14.1$ \\
		& db$16$ & $(5, 6, 8)$ &  $0.978$ $(0.018)$  &  $1.469$ $(0.503)$  &  $\bf{0.252}$ $(0.104)$  &  $\bf{0.254}$ $(0.111)$  & $5.26$ & $\bf{14.0}$ \\
		& db$2$-  & $(2, -, 8)$ & $0.787$ $(0.036)$  &  $13.623$ $(4.868)$  &  $2.968$ $(0.758)$  &  $3.015$ $(0.916)$ & $5.11$ & $14.9$  \\
		& db$4$-  & $(3, -, 8)$ &  $0.873$ $(0.165)$  &  $7.222$ $(6.699)$  &  $1.239$ $(0.569)$  &  $1.213$ $(0.541)$  & $5.11$ & $14.3$    \\
		& db$8$-  & $(4, -, 8)$ &  $0.967$ $(0.013)$  &  $2.215$ $(0.845)$  &  $0.432$ $(0.170)$  &  $0.430$ $(0.177)$  & $5.11$ & $14.0$ \\
		& db$16$- & $(5, -, 8)$  &  $0.977$ $(0.018)$  &  $1.525$ $(0.516)$  &  $0.264$ $(0.104)$  &  $0.264$ $(0.110)$   & $\bf{5.11}$ & $14.0$  \\			
		\hline
		\hline
		\multirow{6}{*}{\rotatebox{90}{Spleen \hphantom{blablabla}}}
		& db1  & $(1, 5, 8)$ &  $0.926$ $(0.034)$  &  $5.203$ $(2.040)$  &  $2.543$ $(0.997)$  &  $2.915$ $(1.276)$ & $9.92$ & $11.6$ \\
		& db2  & $(2, 5, 8)$ &  $0.943$ $(0.030)$  &  $4.200$ $(1.680)$  &  $\bf{1.798}$ $(1.170)$  &  $2.113$ $(1.322)$ & $5.62$ & $10.5$  \\
		& db4  & $(3, 5, 8)$ &  $0.939$ $(0.039)$  &  $4.102$ $(1.637)$  &  $1.904$ $(1.096)$  &  $2.135$ $(1.086)$ & $4.48$ & $9.92$    \\
		& db8  & $(4, 5, 8)$ &  $0.940$ $(0.036)$  &  $4.107$ $(1.572)$  &  $1.821$ $(1.073)$  &  $2.058$ $(1.272)$ & $4.11$ & $9.26$  \\
		& db16 & $(5, 5, 8)$ &  $0.940$ $(0.037)$  &  $4.175$ $(1.371)$  &  $1.880$ $(1.025)$  &  $\bf{1.968}$ $(1.183)$  & $3.98$ & $9.92$  \\
		& db$2$-  & $(2, -, 8)$ & $0.761$ $(0.032)$  &  $14.611$ $(4.810)$  &  $7.344$ $(2.609)$  &  $6.511$ $(1.864)$ & $3.97$ & $9.77$  \\
		& db$4$-  & $(3, -, 8)$ &  $0.930$ $(0.041)$  &  $4.661$ $(1.592)$  &  $2.050$ $(0.768)$  &  $2.429$ $(1.129)$  & $\bf{3.97}$ & $9.35$    \\
		& db$8$-  & $(4, -, 8)$ &  $0.938$ $(0.042)$  &  $4.205$ $(2.303)$  &  $1.990$ $(1.455)$  &  $2.155$ $(1.315)$  & $3.98$ & $8.99$ \\
		& db$16$- & $(5, -, 8)$  &  $0.939$ $(0.035)$  &  $\bf{4.058}$ $(1.517)$  &  $1.798$ $(0.881)$  &  $1.960$ $(1.007)$  & $3.98$ & $\bf{8.79}$  \\			
		& unet & -- &  $\bf{0.952}$ $(0.040)$  &  $4.369$ $(4.913)$  &   -- & -- & $7.76$ & $38.9$  \\
		\hline
		\hline
		\multirow{6}{*}{\rotatebox{90}{Prostate \hphantom{blablabla}}}
		& db$1$  & $(1, 5, 6)$ &  $0.931$ $(0.032)$  &  $5.328$ $(2.587)$  &  $2.369$ $(1.269)$  &  $2.407$ $(0.803)$ & $8.38$ & $11.2$  \\
		& db$2$  & $(2, 5, 6)$ & $0.930$ $(0.035)$  &  $5.450$ $(2.647)$  &  $2.148$ $(1.360)$  &  $2.120$ $(0.788)$  & $5.07$ & $9.66$  \\
		& db$4$  & $(3, 5, 6)$ &  $\bf{0.935}$ $(0.032)$  &  $5.333$ $(2.572)$  &  $\bf{2.026}$ $(1.242)$  &  $\bf{2.005}$ $(0.894)$  & $4.17$ & $8.64$    \\
		& db$8$  & $(4, 5, 6)$ &  $0.931$ $(0.032)$  &  $\bf{5.323}$ $(2.584)$  &  $2.147$ $(1.381)$  &  $2.019$ $(0.873)$  & $3.87$ & $7.59$ \\
		& db$16$ & $(5, 5, 6)$  &  $0.924$ $(0.040)$  &  $5.583$ $(2.665)$  &  $2.218$ $(1.280)$  &  $2.197$ $(0.994)$ & $3.74$ & $7.65$  \\
		& db$2$-  & $(2, -, 6)$ & $0.779$ $(0.025)$  &  $14.584$ $(4.487)$  &  $6.072$ $(1.937)$  &  $6.577$ $(1.998)$ & $3.73$ & $8.22$  \\
		& db$4$-  & $(3, -, 6)$ &  $0.921$ $(0.037)$  &  $6.040$ $(3.264)$  &  $2.353$ $(1.316)$  &  $2.328$ $(1.046)$  & $3.73$ & $7.84$    \\
		& db$8$-  & $(4, -, 6)$ &  $0.923$ $(0.041)$  &  $6.082$ $(3.380)$  &  $2.334$ $(1.485)$  &  $2.328$ $(1.008)$  & $\bf{3.73}$ & $7.79$ \\
		& db$16$- & $(5, -, 6)$  &  $0.928$ $(0.032)$  &  $5.499$ $(2.901)$  &  $2.177$ $(1.273)$  &  $2.078$ $(0.784)$  & $3.74$ & $\bf{7.56}$  \\	
		& unet & -- &  $0.932$ $(0.047)$  &  $5.673$ $(2.402)$  &  -- & -- & $7.76$ & $37.2$ \\
		\hline					
		\bottomrule
	\end{tabular}
	}
	\label{table:results}
\end{table}

\paragraph{Spleen}
The predictions of our models are accurate and on par with the baseline U-Net. In terms of the dice score, 
the U-Net performs slightly better. This is mostly due to ``edge'' cases where the boundary of the 
spleen is small and about to disappear from our two-dimensional sliced-view. In such cases it may sometimes
be ambiguous to define an accurate ground-truth contour.  On the other hand, our models perform slightly better 
in terms of Hausdorff-distance. This is reflected in smaller averages, but more noticeably our models 
follow the boundaries more faithfully resulting in a smaller variation around the mean. 

\paragraph{Prostate}
The wavelet models produce accurate 
predictions and are on par with the U-Net. Our models perform slightly better in terms of 
Hausdorff distance. While our models produce accurate predictions for most examples, there are instances where
both our model and the baseline U-Net fail to produce accurate predictions; see the
last two columns of Figure \ref{fig:example_pred}. In these examples, the detail 
coefficients associated to parts of the curve with high curvature are too small in magnitude 
to be accurately predicted. While detail coefficients of such small magnitude were less relevant in the latter two examples,
they are important for the prostate. 

\paragraph{Ablation study (no detail coefficients)}
To demonstrate the importance of the detail coefficients, we have trained models without them (and skip-connections). 
In this set up, we do not supervise the predictions on the lowest resolution level during training.  
The results demonstrate, as expected, that the lower-order wavelets db$1$, db$2$ perform significantly worse without detail coefficients.
In fact, for db$1$ our model failed to produce any sensible approximations that can be evaluated and were therefore omitted.
A small drop in performance is observed for db$4$.  
In general, for db$16$ there is not much gain, with respect to accuracy, memory-footprint and inference time, to 
explicitly model the detail coefficients.

\section{Conclusion}
\label{sec:conclusion}

We have introduced a novel method to model boundaries of two dimensional simply connected domains using wavelets and MRAs. 
In effect this allows for subpixel segmentations. The efficacy of the method has been demonstrated by modeling the boundaries 
of hypercloids (toy-example), spleen and prostate, and demonstrating that the results are on par with a traditional U-Net
yielding up to five times faster inference speed.

\nocite{*}
\bibliographystyle{spbasic}
\bibliography{Bibliography}

\appendix
\newpage
\section{Theoretical results}
\label{sec:appendix}

In this section we prove theoretical results regarding the computation of 
wavelet decompositions. In addition, we provide all the implementation details. 

\subsection{Explicit expression for $a_{jk}$}
\label{appendix: init_approx_coeffs}
To address the issue that periodic signals are not contained in $L^{2}(\RR)$, we consider the
cut-off $\tilde \gamma := \gamma \bold{1}_{[-l, l]}$.
In this section we derive an explicit formula for $\langle \tilde \gamma, \varphi_{jk} \rangle$ by exploiting
the periodicity of $\gamma$. 
 In particular, we will quantify the claim that (scaled) sample values of $\gamma$ 
may be used as approximation coefficients. For this purpose, we first review some facts about the scaling equation. 
If we take the Fourier transform of both sides of the scaling equation, we obtain the relation
$\hat \varphi (\xi) = H(\frac{\xi}{2}) \hat \varphi(\frac{\xi}{2})$ for $\xi \in \RR$, where $H(\xi) = \frac{1}{\sqrt{2}}\sum_{k \in \ZZ} h_{k} e^{-i 2 \pi \xi k}$. 
The map $H$ is referred to as the \emph{refinement mask} associated to $h$. 
Throughout this paper we assume that $h$ is finite, in which case $\varphi$ has compact support and 
$H$ is a trigonometric polynomial of period $1$. It is beyond the scope of this paper to discuss the properties of 
$H$ in detail and refer the reader to \cite{HarmonicAnalysis, WaveletsTheory}. Here we only need the following result. 
Under suitable conditions, one may iterate the equation for $\hat \varphi$
and show that $\hat \varphi(\xi) = \prod_{k=1}^{\infty} H( \frac{\xi}{2^{k}})$ for all $\xi \in \RR$.

\begin{lemma}[Initialization approximation coefficients]
	\label{lemma:approx_coeffs}
	Let $h \in \elltwo$ be a low-pass-filter defining a MRA on $\Ltwo$ and $H$ the associated refinement mask.
	Assume $h$ is non-zero for only a finite number of coefficients, $\mbox{supp}(\varphi) \subset [0, \beta]$ for some $\beta >0$
	and $\varphi$ is continuous. 
	Furthermore, suppose that $\hat \varphi(\xi) = \prod_{n=1}^{\infty} H( \frac{\xi}{2^{n}})$ for all $\xi \in \RR$. 
	If $\gamma \in C_{\text{per}}^{2}([0, l])$ is a $l$-periodic map with Fourier coefficients $\left( \gamma_{m} \right)_{m \in \ZZ}$, 
	then 
	\begin{align}
		\label{eq:approx_coeffs}
		\left \langle \tilde \gamma, \varphi_{jk} \right \rangle = 
		2^{-\frac{j}{2}} \sum_{m \in \ZZ} \gamma_{m} e^{i \omega(l) m \frac{k}{2^{j}} } \prod_{n=1}^{\infty} H \left( -\frac{m}{ l 2^{j+n}} \right),
	\end{align}
	where $\omega(l) := \frac{2 \pi}{l}$ is the angular frequency of $\gamma$, 
	for any $j \in \ZZ$ and $k \in \{\lceil -2^{j}l \rceil, \ldots, \lfloor 2^{j} l - \beta \rfloor \}$. 
	\begin{proof}
		Let $j \in \ZZ$ and $k \in \{\lceil -2^{j}l \rceil, \ldots, \lfloor 2^{j} l - \beta \rfloor \}$ be arbitrary. 
		A change of variables shows that
		\begin{align*}
			\left \langle \tilde \gamma, \varphi_{jk} \right \rangle = 
			2^{-\frac{j}{2}} \int_{[0, \beta]} \tilde \gamma \left( 2^{-j}(t+k) \right) \varphi(t) \dt, \quad k \in \ZZ,
		\end{align*}
		since  $\mbox{supp}(\varphi)\subset [0, \beta]$. 
		In particular, note that $2^{-j}(t+k) \in [-l, l]$ for $t \in [0, \beta]$, since $k \in \{\lceil -2^{j}l \rceil, \ldots, \lfloor 2^{j} l - \beta \rfloor \}$. 
		Therefore, we may plug in the Fourier expansion for $\gamma$ and compute
		\begin{align*}
			\int_{[0, \beta]} \tilde \gamma \left( 2^{-j}(t+k) \right) \varphi(t) \dt = 
			 \int_{[0, \beta]} \sum_{m \in \ZZ} \gamma_{m} e^{i \omega(l) m \frac{t + k }{2^{j}}} \varphi(t) \dt.
		\end{align*}
		Next, note that that series inside the integral converges pointwise to $\gamma\left(2^{-j}(t+k)\right) \varphi(t)$ on $[0, \beta]$.
		Furthermore, the partial sums can be bounded from above on $[0, \beta]$ by a constant, since $\gamma \in C_{\text{per}}^{2}([0, l])$ and $\varphi$ is bounded. 
		Therefore, we may interchange the order of summation and integration by the Dominated Convergence Theorem:
		\begin{align*}
			\int_{[0, \beta]} \sum_{m \in \ZZ} \gamma_{m} e^{i \omega(l) m \frac{t + k }{2^{j}}} \varphi(t) \dt &= 
			\sum_{m \in \ZZ} \gamma_{m} e^{i \omega(l) m \frac{k}{2^{j}} } \int_{[0, \beta]} e^{i \omega(l) m \frac{t}{2^{j}} } \varphi(t) \dt.
		\end{align*}
		Finally, changing the domain of integration to $\RR$ again, we see that
		\begin{align*}
			\sum_{m \in \ZZ} \gamma_{m} e^{i \omega(l) m \frac{k}{2^{j}} } \int_{[0, \beta]} e^{i \omega(l) m \frac{t}{2^{j}} } \varphi(t) \dt = 
			\sum_{m \in \ZZ} \gamma_{m} e^{i \omega(l) m \frac{k}{2^{j}} } \hat \varphi \left( - \frac{m}{l2^{j}} \right).
		\end{align*}
		The stated result now follows from the assumption that $\hat \varphi(\xi) = \prod_{k=1}^{\infty} H( \frac{\xi}{2^{k}})$ for any $\xi \in \RR$. 
	\end{proof}
\end{lemma} 

\begin{remark}
The bounds $\lceil -2^{j}l \rceil$ and $\lfloor 2^{j} l - \beta \rfloor$ are the smallest and largest integer, respectively, for which
$2^{-j}(t+k) \in [-l, l]$ for all $t \in [0, \beta]$. The bounds on $k$ are somewhat artificial, however, since the argument may be repeated
for any truncation of $\gamma$ on $[-nl, nl]$, where $n \in \NN$. This observation is reflected in the righthand-side of \eqref{eq:approx_coeffs},
which is well-defined for all $k \in \ZZ$ (but not squared-summable).  
The reason for choosing this particular truncation is to identify a minimal number of approximation coefficients
needed to cover the full signal $\gamma$.
\end{remark}

The requirement that $\hat \varphi(\xi) = \prod_{k=1}^{\infty} H( \frac{\xi}{2^{k}})$ for $\xi \in \RR$ is satisfied for all MRA's 
considered in this paper. In particular, we remark that $\hat \varphi$ is 
continuous at $\xi=0$ and $\hat \varphi(0) = 1$. It follows from these observations and Lemma \ref{lemma:approx_coeffs} that 
$a_{jk} (\tilde \gamma) =  a_{jk} (\gamma) \approx 2^{-\frac{j}{2}} \gamma( k 2^{-j} )$ for $j$ sufficiently large and $k$ constrained
to  $\{\lceil -2^{j}l \rceil, \ldots, \lfloor 2^{j} l - \beta \rfloor \}$.	
	
\subsection{Pyramid Algorithm - implementation details}
\label{appendix:implementation}
In this section we provide the computational details for how to compute approximation and detail coefficients using the Discrete Fourier Transform (DFT). 
In the following arguments all sequences are assumed to be two-sided. We will frequently abuse notation and write that a 
finite sequence is an element in $\CC^{n}$ or $\RR^{n}$. What we actually mean by this, is that we have a two-sided sequence
of length $n$ which can be embedded in $\ell^{2}(\ZZ)$ by appropriately padding with zeros. In such a situation we will explicitly specify the indices of 
the sequence so that the intended ordering is clear. Similarly, all operators in this section are implicitly assumed to be defined on the 
associated (two-sided) sequence spaces, even though we may write that they are defined on $\CC^{n}$ or $\RR^{n}$. 

Let $a = (a_{k})_{k=-N}^{N-1} \in \RR^{2N}$ and $g = (g_{k})_{k=1-M}^{M-1} \in \RR^{2M -1}$ be arbitrary. Here $g$ 
may be interpreted as a high-pass-filter of length $M$ with $N \geq M \geq 2$. Similarly, we may think of $a$ as the approximation coefficients 
of a $1$-periodic signal at a specific resolution level $j$, with $N = 2^{j-1}$, as explained in Section \ref{sec:init_approx_coeffs}. 
In this section, however, we will not emphasize these interpretations, e.g., write $a_{jk}$ instead of $a_{k}$, to 
avoid clutter in the notation. Observe that the righthand-side of \eqref{eq:approx_coeffs} is a $2^{j}$-periodic sequence 
for $1$-periodic signals. Therefore, to properly deal with ``boundary terms'', we will use 
the $2N$-periodic extension $\tilde a \in \RR^{\ZZ}$ of $a$ to evaluate discrete convolutions. 

\paragraph{Convolution and multiplication of trigonometric polynomials}
To compute the detail and approximation coefficients, we need to evaluate expressions of the form 
\begin{align}
	\label{eq:circ_conv}
	(\tilde a \ast g)_{k} = \sum_{\substack{k_{1} + k_{2} = k \\ \vert k_{2} \vert \leq M-1 \\ k_{1} \in \ZZ}} \tilde a_{k_{1}} g_{k_{2}}
	 = \sum_{\vert k_{2} \vert \leq M - 1} \tilde a_{k - k_{2}} g_{k_{2}}
\end{align}
for $- N \leq k \leq N-1$. Note that although $\tilde a$ is an infinite $2N$-periodic sequence, 
the series in \eqref{eq:circ_conv} contains only a finite number of nonzero-terms, since $g_{k_{2}} =0$ for $\vert k_{2} \vert \geq M$. 
Furthermore, for $- N \leq k \leq N-1$, we do not need the full periodic extension $\tilde a$, but only a partial (finite) extension 
$P_{K}(a)$, where $P_{K} : \CC^{2N} \rightarrow \CC^{K}$ is defined by 
\begin{align*}
	(P_{K}(a))_{k} := 
	\begin{cases}
		a_{k + 2N} & 1 - N -M \leq k \leq -N -1, \\[2ex]
		a_{k} & -N \leq k \leq N-1, \\[2ex]
		a_{k - 2N} & N \leq k \leq N + M -2
	\end{cases}
\end{align*}
and $K := 2(N+M -1)$. 
That is, $(\tilde a \ast g)_{k} = (P_{K}(a) \ast g)_{k}$ for $- N \leq k \leq N-1$. 

We use standard arguments to compute $P_{K}(a) \ast g$. Namely, we interpret $P_{K}(a) \ast g$ as the
Fourier coefficients of $uv$, where $u, v: \RR \rightarrow \CC$ are the trigonometric polynomials defined by 
\begin{align*}
	u(\theta) = \sum_{k=-\frac{K}{2}}^{\frac{K}{2} - 1} (P_{K}(a))_{k} e^{ik\theta}, \quad v(\theta) = \sum_{k=1-M}^{M - 1} g_{k} e^{ik\theta}.
\end{align*}
The product $uv$ is a trigonometric polynomial with $\tilde K := K + 2(M-1)$ non-zero coefficients corresponding to terms of order
$-\frac{\tilde K}{2} \leq k \leq \frac{\tilde K}{2} - 1$. The coefficients of $uv$ can be characterized by evaluating it
on $\tilde K$ distinct points in $\CC$. After fixing $\tilde K$ such points, we may go back and forth between value and coefficient representations
of $u$, $v$ and $uv$ using the isomorphism defined by the evaluation operator. 

\paragraph{Evaluation at the roots of unity}
We evaluate $u$ and $v$, in the complex variable $z = e^{i \theta}$, at the $\tilde K$-th roots of unity. To do this, we first extend $P_{K}(a)$ and $g$ to 
sequences in $\CC^{\tilde K}$ by padding with zeros. More precisely, define
$Z^{\text{even}}_{\tilde K} : \CC^{K} \rightarrow \CC^{\tilde K}$ by 
$(Z^{\text{even}}_{\tilde K}(b))_{k} = 	b_{k}$ for $-\frac{K}{2} \leq k \leq \frac{K}{2} - 1$ and zero for 
$-\frac{\tilde K}{2} \leq k < - \frac{K}{2}$ and $\frac{K}{2} \leq k < \frac{\tilde K}{2}$. Similarly, define 
$Z^{\text{odd}}_{M}: \CC^{2M -1} \rightarrow \CC^{\tilde K}$ by 
$(Z^{\text{odd}}_{M}(b))_{k} = b_{k}$ for $\vert k \vert < M$ and zero for 
$-\frac{\tilde K}{2} \leq k \leq -M$ and $M \leq k < \frac{\tilde K}{2}$.
We can now evaluate $u$ and $v$ at the $\tilde K$-th roots of unity by computing
${\bf{DFT}}_{\tilde K} \circ S_{\tilde K} \circ Z^{\text{even}}_{\tilde K}   \circ P_{K}(a)$ 
and ${\bf{\hat G}}_{\tilde K} := {\bf{DFT}}_{\tilde K} \circ S_{\tilde K} \circ Z^{\text{odd}}_{M}(g)$, respectively, where 
$S_{\tilde K} : \CC^{\tilde K} \rightarrow \CC^{\tilde K}$ is defined by
\begin{align*}
	(S_{\tilde K}b)_{k} := 
	\begin{cases}
		b_{k}, & 0 \leq k \leq \frac{\tilde K}{2} - 1, \\[2ex]
		b_{k - \tilde K}, & \frac{\tilde K}{2} \leq k \leq \tilde K - 1.
	\end{cases}
\end{align*}
Consequently, $uv$ can be evaluated at the $\tilde K$-th roots of unity by taking the element-wise product of the latter two vectors. 
Finally, the desired coefficients $(P_{K}(a) \ast g)_{k=-N}^{N-1}$ are obtained by going back to coefficient
space using the inverse DFT , i.e., 
\begin{align*}
	\label{eq:trunc_conv}
	(P_{K}(a) \ast g)_{k=-N}^{N-1} = 
	\Pi_{N}
	S_{\tilde K}^{-1} {\bf{DFT}}_{\tilde K}^{-1} 
	\left( 
		{\bf{\hat G}}_{\tilde K} 
		\ \odot \ 
		{\bf{DFT}}_{\tilde K} \circ S_{\tilde K} \circ Z^{\text{even}}_{\tilde K}   \circ P_{K}(a)
	\right).
\end{align*}
Here  $\odot$ denotes the Hadamard-product and $\Pi_{N} : \CC^{\tilde K} \rightarrow \CC^{2N}$ is the truncation operator defined by 
$\Pi_{N}(b) := (b_{k})_{k=-N}^{N-1}$. 
	
\section{Preprocessing}
In this section we provide the details of our preprocessing steps. 

\subsection{Truncation Fourier coefficients}
\label{appendix:truncation}
The magnitude of the approximated Fourier coefficients will typically stagnate and stay constant (approximately) beyond some critical order, 
since all computations are performed in finite (single) precision. We locate this critical order $m^{\ast}_{0}(s) \in \NN$ for each component $s \in \{1, 2\}$, 
if present, by iteratively fitting the best line, in the least squares sense, through the points 
$\left \{ \left(m,  \left \Vert \left( \left \vert \left[ \tilde \gamma_{\tilde m} \right]_{s} \right \vert \right)_{\tilde m=m_{0}}^{m} \right \Vert_{1} \right) : m_{0} \leq m \leq N -1 \right \}$ 
for $1 \leq m_{0} \leq N-1$. We iterate this process until the residual is below a prescribed threshold $\delta_{N} >0$. In practice, 
we set $\delta_{N} = 0.1$. The Fourier coefficients with index strictly larger than $m_{0}^{\ast}(s)$ are set to zero. 

\subsection{Consistent parameterizations}
\label{appendix:midpoint}
To have consistent parameterizations we enforce that all contours start at angle zero at time zero relative to the midpoint 
$c = (c_{1}, c_{2} ) \in \RR^{2}$ of the region of interest $R$. This is accomplished by exploiting the Fourier representation of the curve. 
More precisely, let $\tilde \gamma := t \mapsto \sum_{\vert m \vert \leq N-1} \tilde \gamma_{m} e^{i \omega(l)m t}$, where $\omega(l) = \frac{2 \pi}{l}$, 
be the contour with Fourier coefficients $\Gamma :=(\tilde \gamma_{m})_{m = 1-N}^{N-1}$.
The midpoint $c$ of the region enclosed by $\tilde \gamma$ is given by 
\begin{align}
	c_{s}  &= \frac{1}{ \lambda(R)} \int_{R} u_{s} \ \mbox{d} \lambda(u_{1}, u_{2}) 
	= (-1)^{s} \frac{ \left(  [\Gamma]_{1} \ast
	 [\Gamma]_{2} \ast [ \Gamma']_{s} \right)_{0} }{  \left( [\Gamma]_{1} \ast [ \Gamma' ]_{2} \right)_{0}},
	\quad s \in \left \{1, 2  \right \}
\end{align}
by Green's Theorem. Here $\lambda$ denotes the Lebesgue measure on $\RR^{2}$, $[\Gamma]_{s}$ are the Fourier coefficients of $[\tilde \gamma]_{s}$, 
and $(\Gamma')_{m} := i m \omega(l) \gamma_{m}$ for $\vert m \vert \leq N-1$. We can now compute the desired parameterization by determining
$\tau \in [0, l]$ such that $\text{arccos} \left( \dfrac{[ \tilde \gamma (- \tau) - c ]_{1}}{ \Vert \tilde \gamma(-\tau) - c \Vert_{2}} \right) \approx 0$ and 
defining $\gamma (t):= \tilde \gamma(t - \tau)$. 
While $\tau$ can be easily found using Newton's method, it suffices in practice to simply re-order $y$ from the start, before computing the Fourier coefficients.
More precisely, we first define a shift $\tilde y$ of $y$ by $\tilde y_{k} := y_{k \ + \ k^{\ast} \ \text{mod} \ n_{p}}$ for $0 \leq k \leq n_{p} -1$, 
where $k^{\ast} := \text{argmin} \left \{ \text{arccos} \left( \dfrac{[ y_{k} - c ]_{1}}{ \Vert y_{k} - c\Vert_{2}} \right) \right \}_{k = 0}^{n_{p} - 1}$. 

\newpage 
\section{Figures}	
\label{appendix:visualization}
In this section we provide additional visualizations of the statistics, wavelet decompositions and predicted contours. 

\subsection{Toy example}
\begin{figure}[htb]
	\centering
	\subfloat[\centering Dice]{{\includegraphics[width=0.49\textwidth]{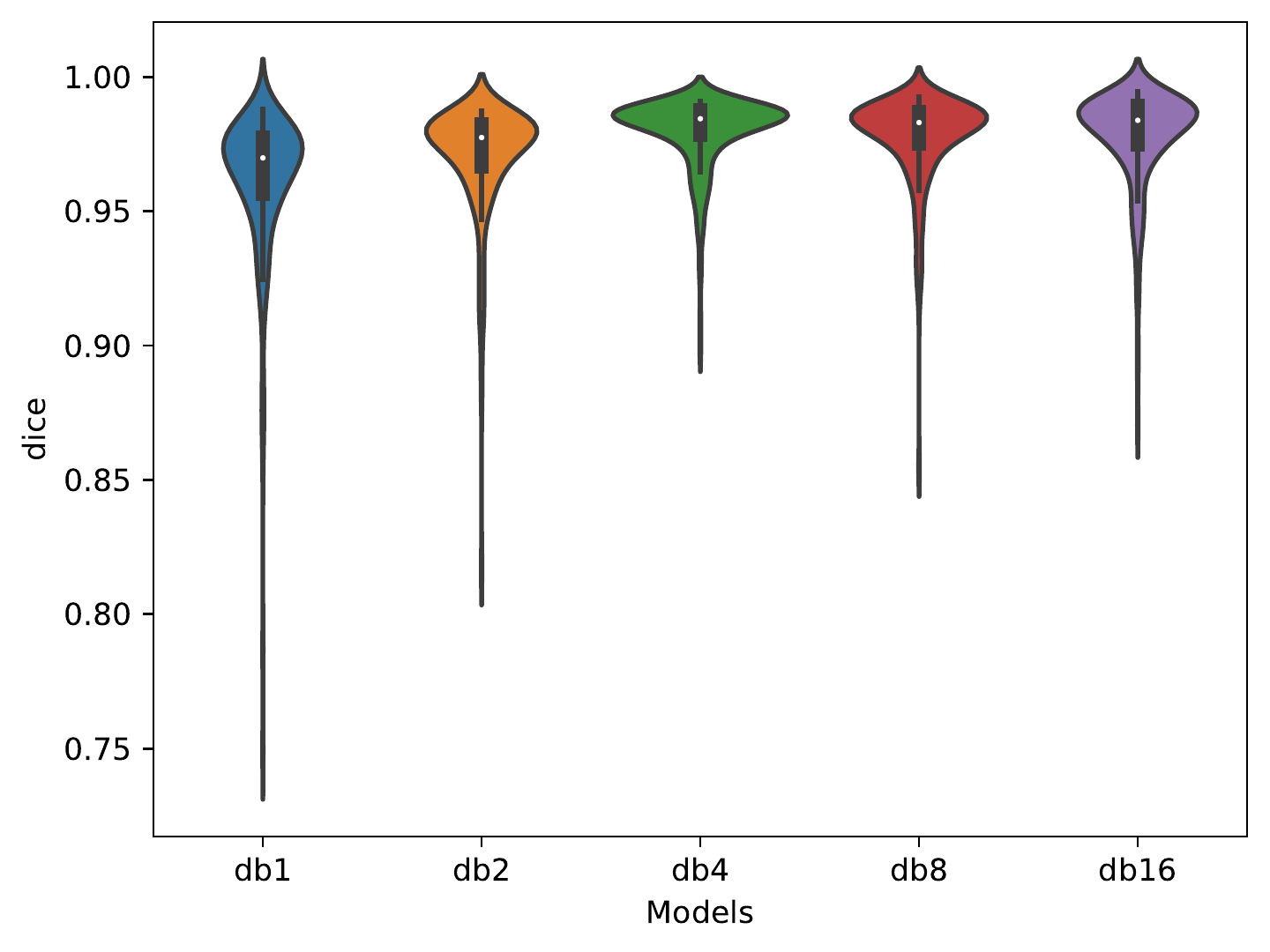}}}
	\subfloat[\centering Hausdorff]{{\includegraphics[width=0.48\textwidth]{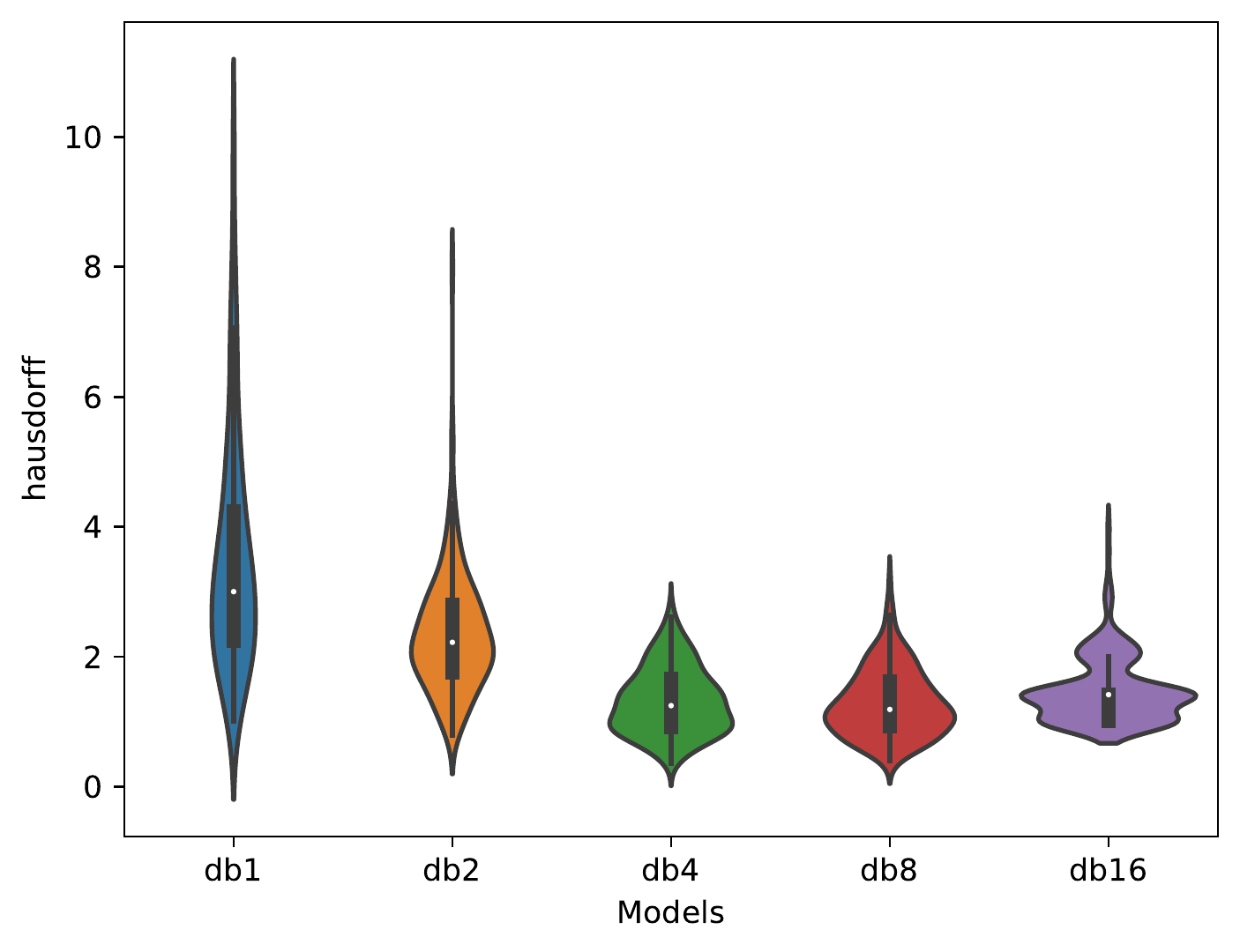}}} \\
	\caption{Boxplots and visualization of approximate densities for the dice scores and Hausdorff distances for the
	toy problem.}
	\label{fig:toy_boxplot}
\end{figure}

\begin{figure}[!htb]
	\centering
	\subfloat[]{{\includegraphics[width=0.97\textwidth]{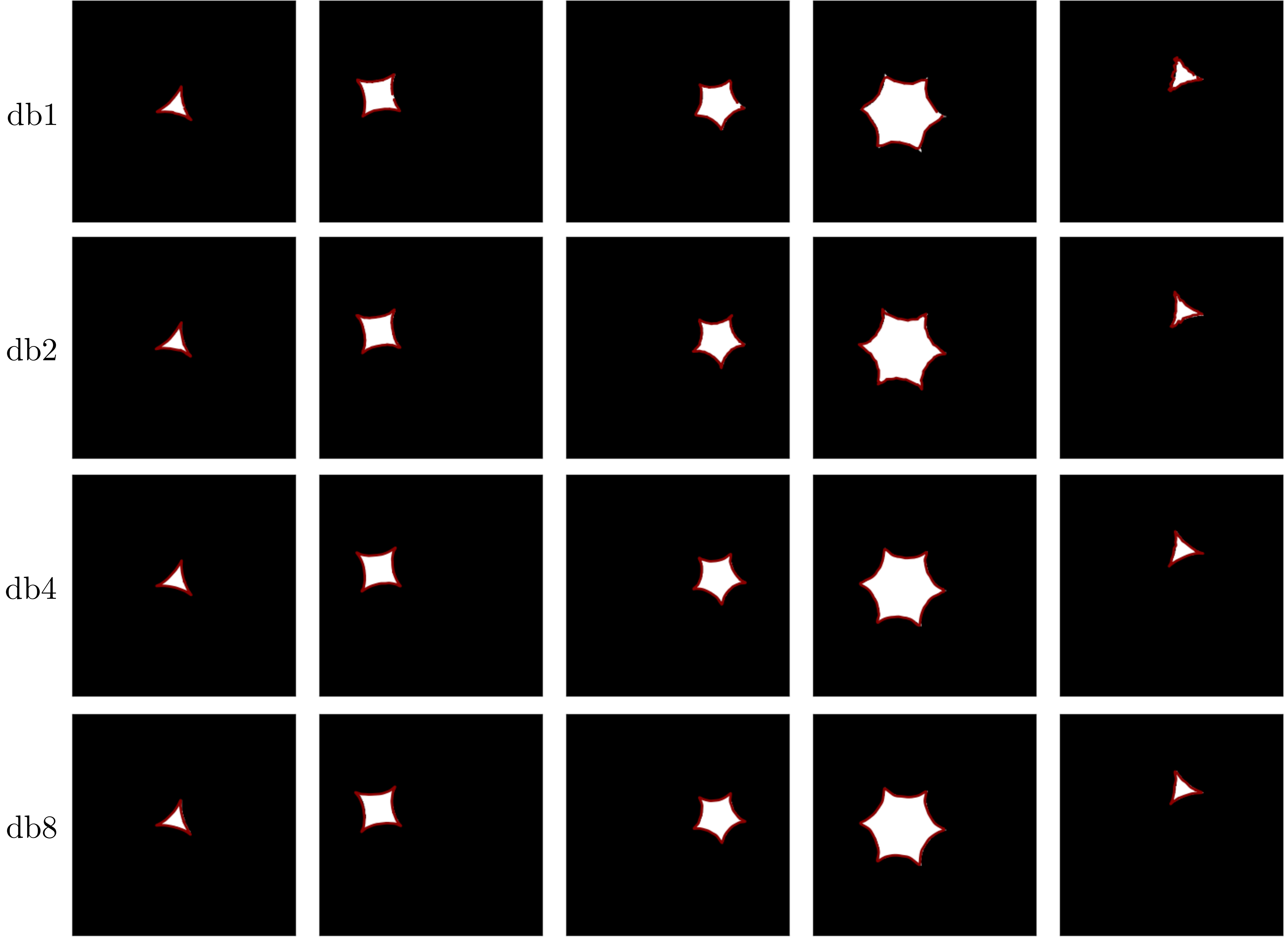}}}
	\caption{Predicted curve (in red) for a binary mask (in white). In order to visualize
	the predicted curve without too much clutter we have not depicted the ground-truth contour. }
	\label{fig:toy_pred}
\end{figure}

\begin{figure}[!htbp]
	\centering
	\subfloat[db$1$]{{\includegraphics[width=0.5\textwidth]{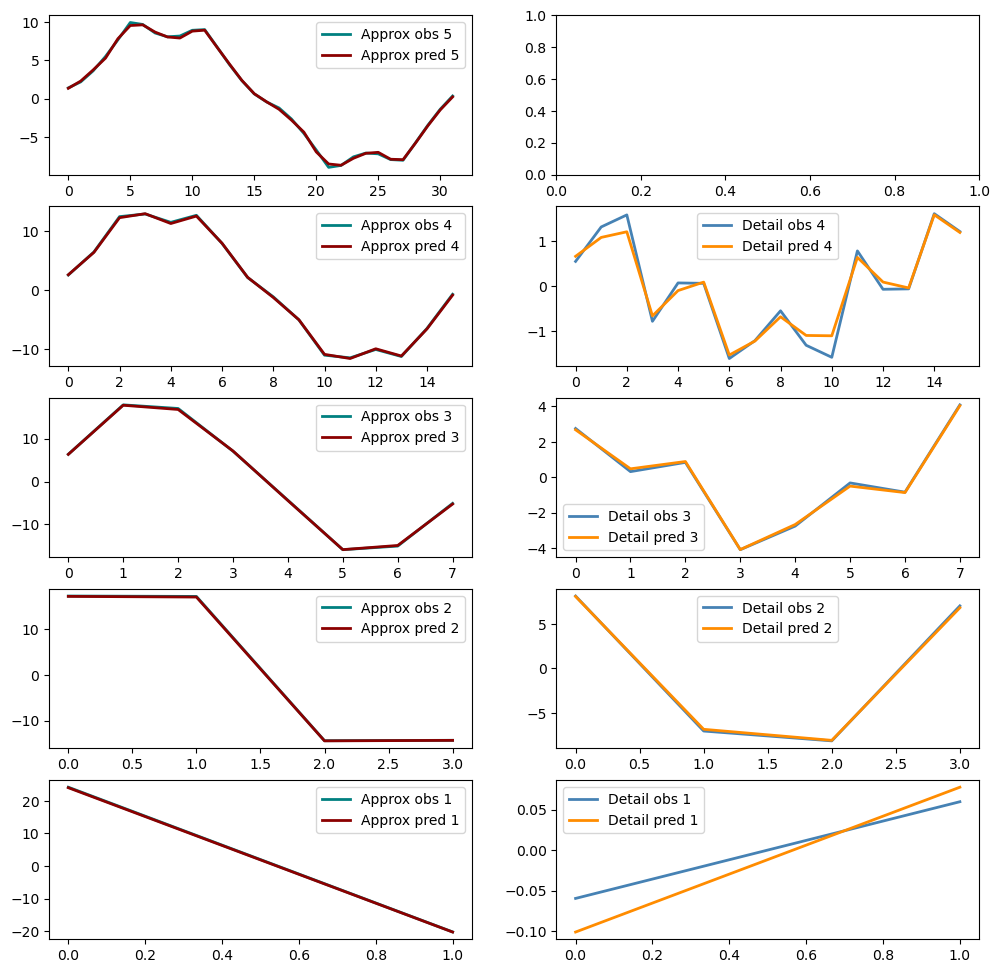}}}
	\subfloat[db$2$]{{\includegraphics[width=0.5\textwidth]{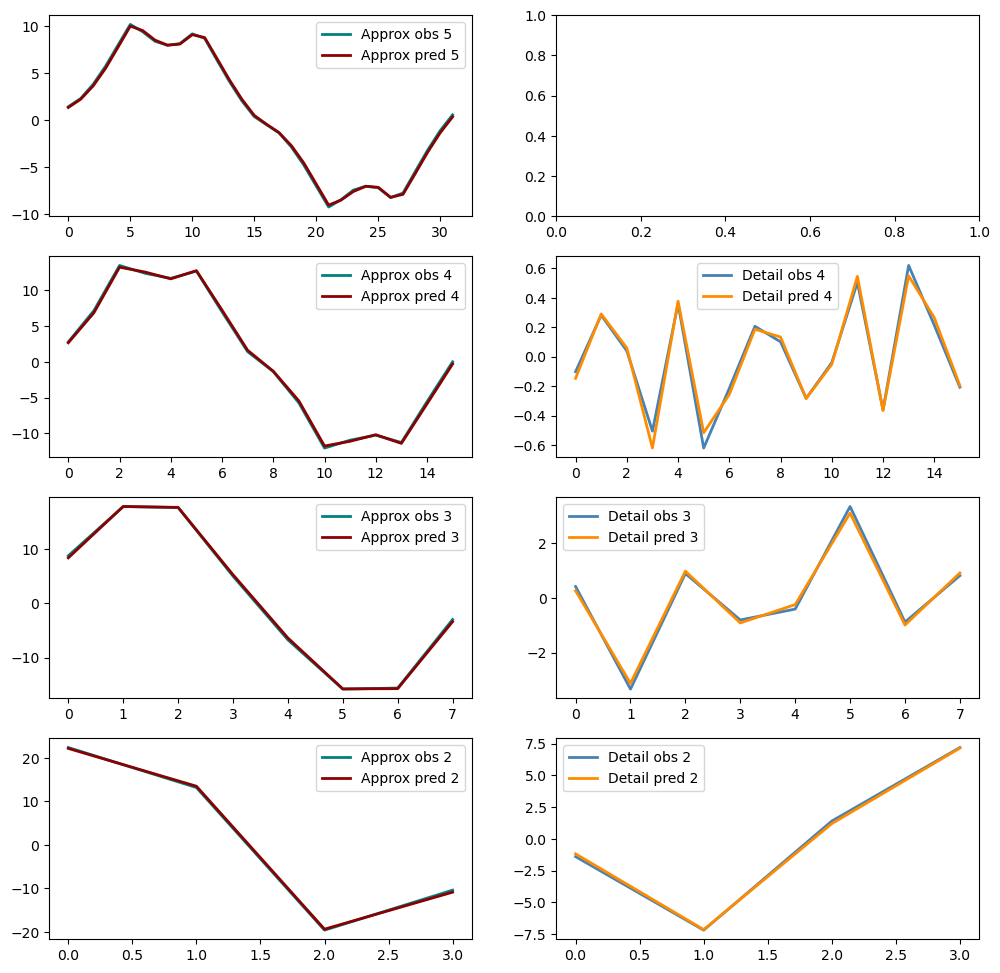}}} \\
	\subfloat[db$4$]{{\includegraphics[width=0.5\textwidth]{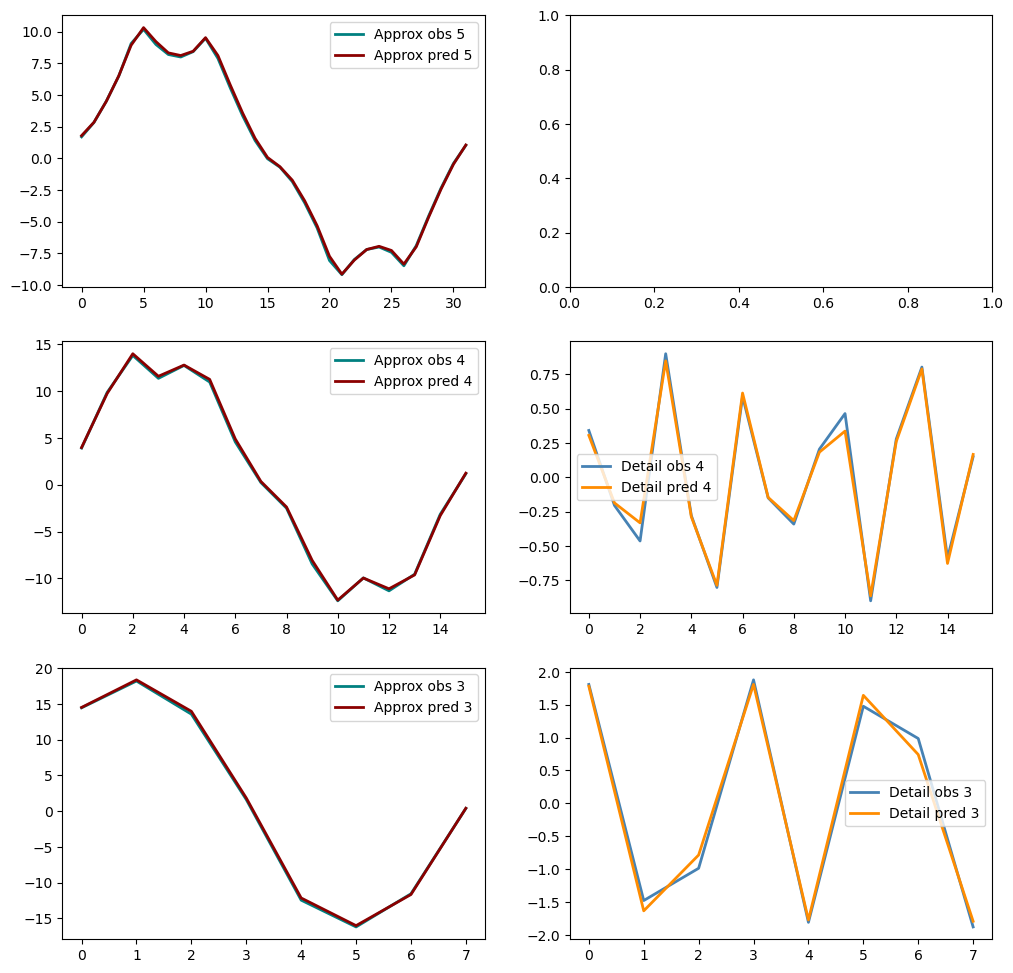}}}
	\subfloat[db$8$]{{\includegraphics[width=0.5\textwidth]{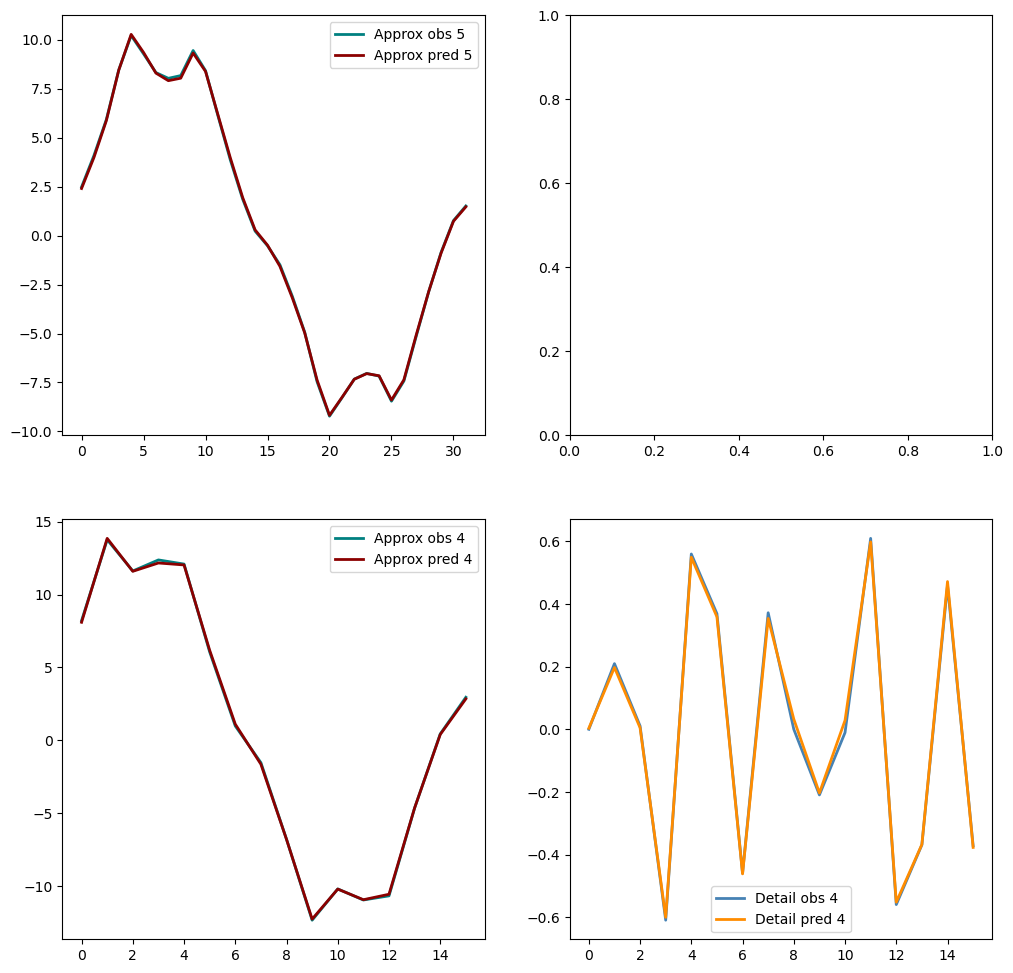}}}	
	\caption{Predicted and observed wavelet decompositions of the first component of the contour depicted in the fourth column in 
	Figure \ref{fig:toy_pred}.}
	\label{fig:toy_wav}
\end{figure}

\newpage
\subsubsection{Spleen}
\begin{figure}[htb]
	\centering
	\subfloat[\centering Dice]{{\includegraphics[width=0.49\textwidth]{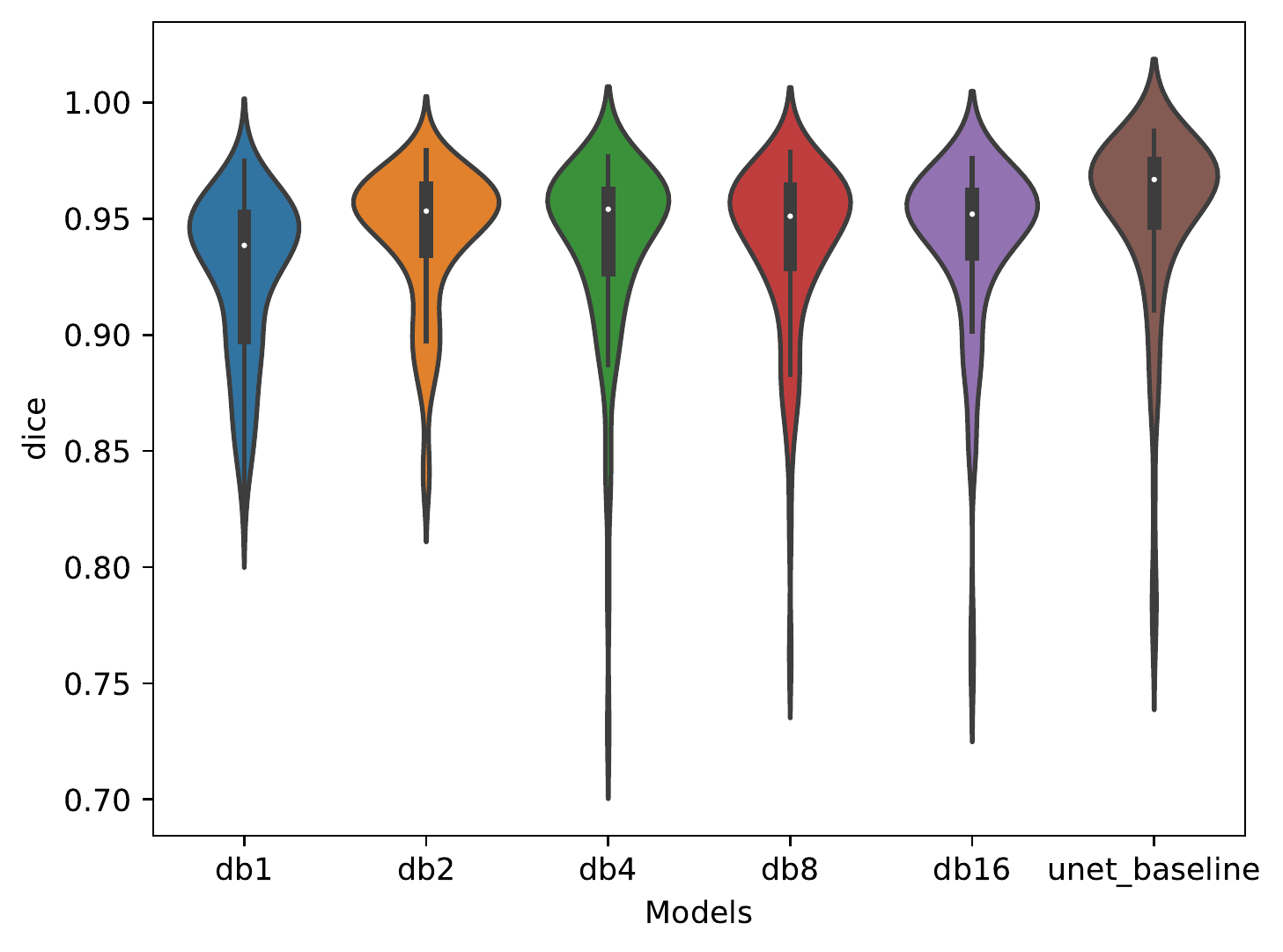}}}
	\subfloat[\centering Hausdorff]{{\includegraphics[width=0.48\textwidth]{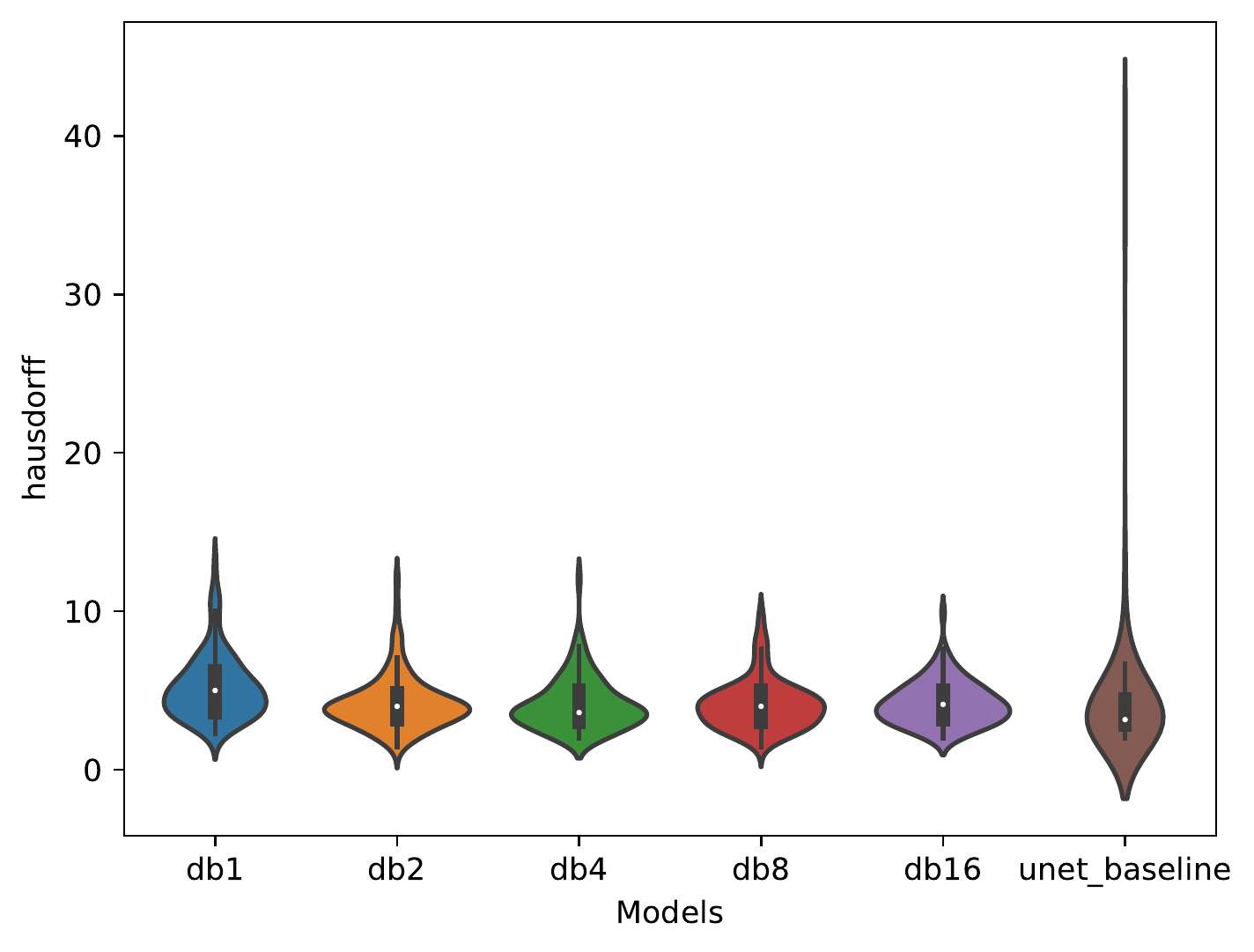}}} \\
	\caption{Boxplots and visualization of approximate densities for the dice scores and Hausdorff distances for the
	spleen.}
	\label{fig:toy_boxplot}
\end{figure}
\begin{figure}[!htb]
	\centering
	\subfloat[]{{\includegraphics[width=0.97\textwidth]{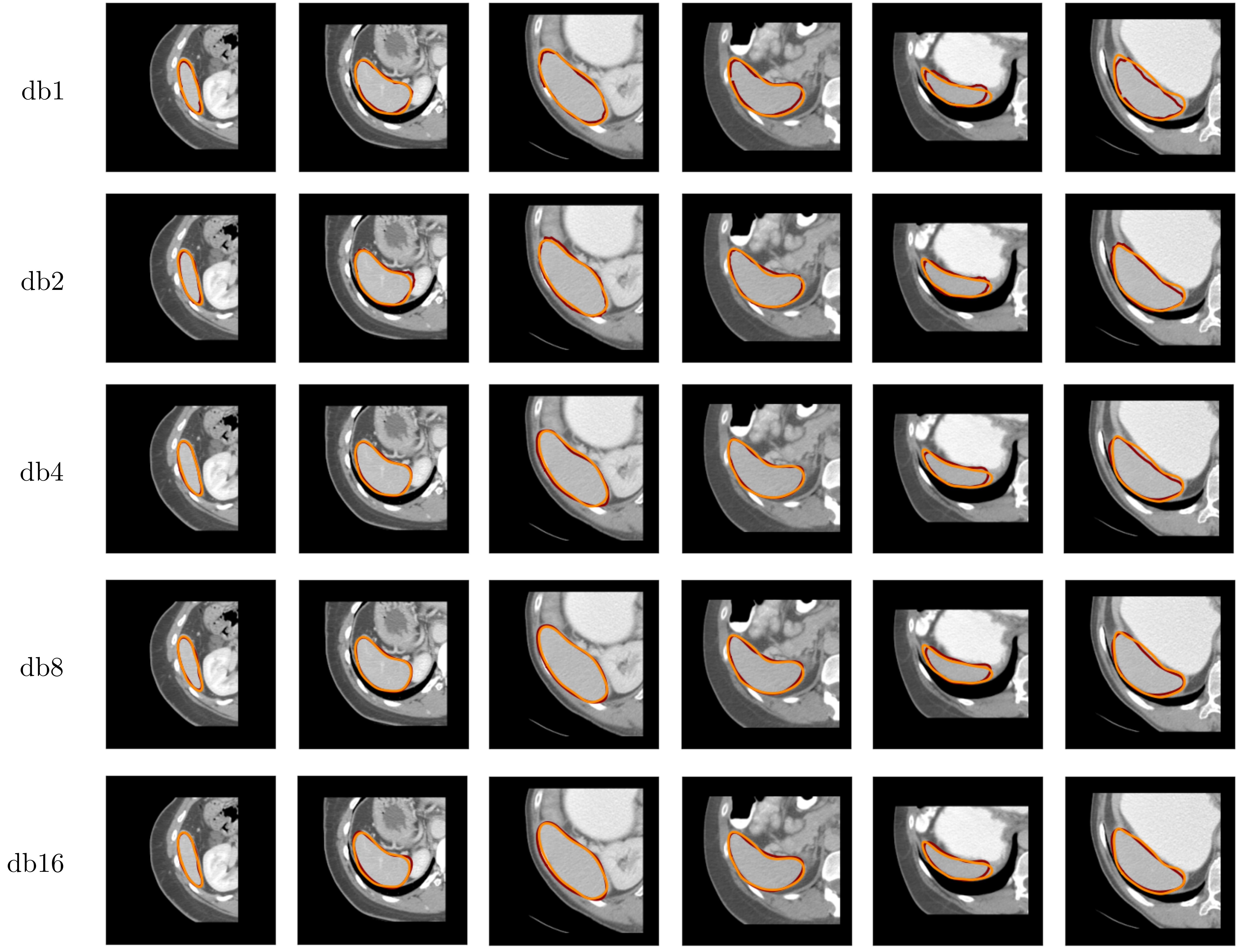}}}
	\caption{Predicted and observed boundaries colored in red and orange, respectively, of the spleen. The
		      last two columns correspond to a few ``hard'' examples.}
	\label{fig:spleen_pred}
\end{figure}

\begin{figure}[!htbp]
	\centering
	\subfloat[db$1$]{{\includegraphics[width=0.5\textwidth]{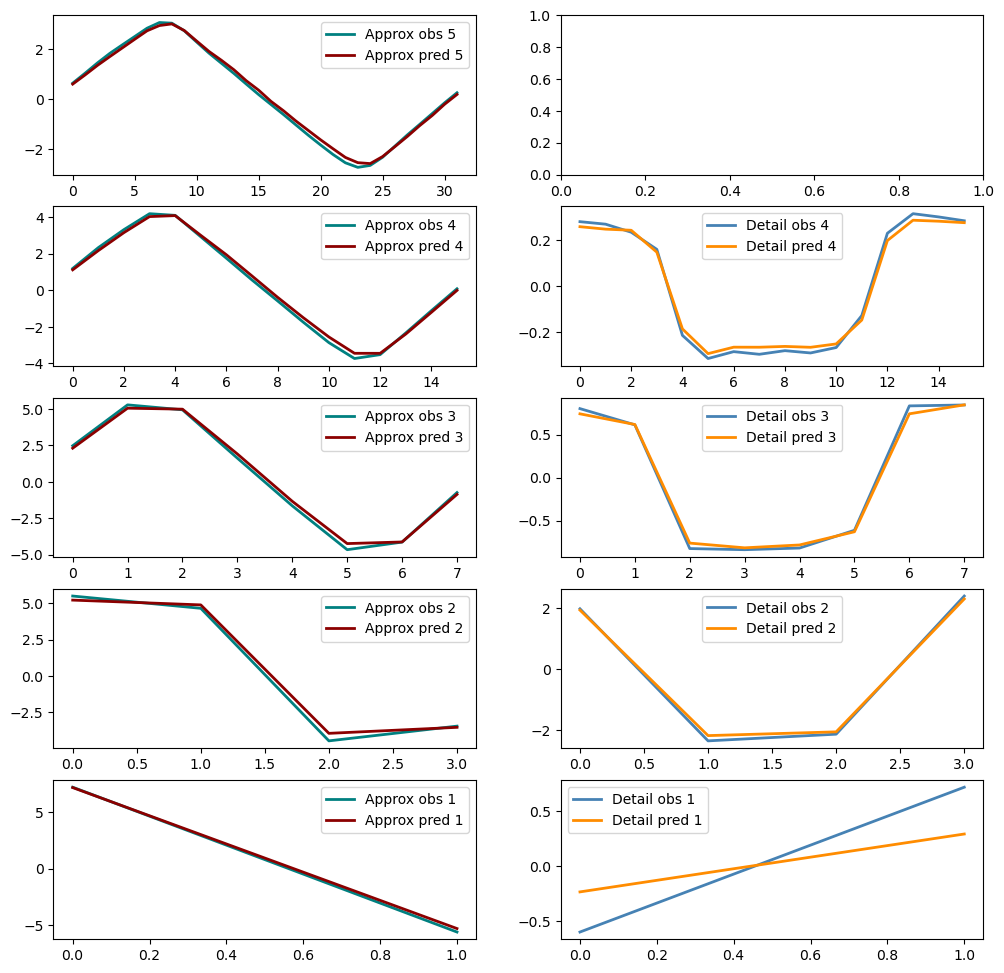}}}
	\subfloat[db$2$]{{\includegraphics[width=0.5\textwidth]{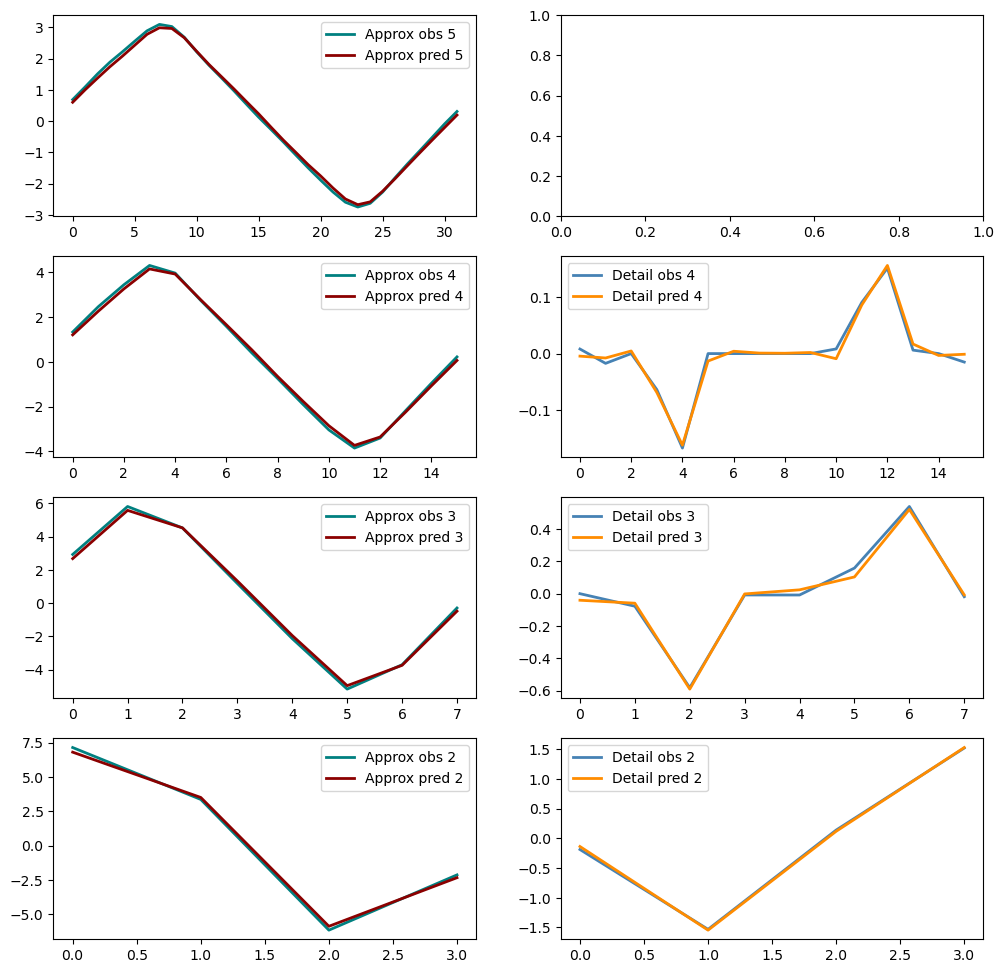}}} \\
	\subfloat[db$4$]{{\includegraphics[width=0.5\textwidth]{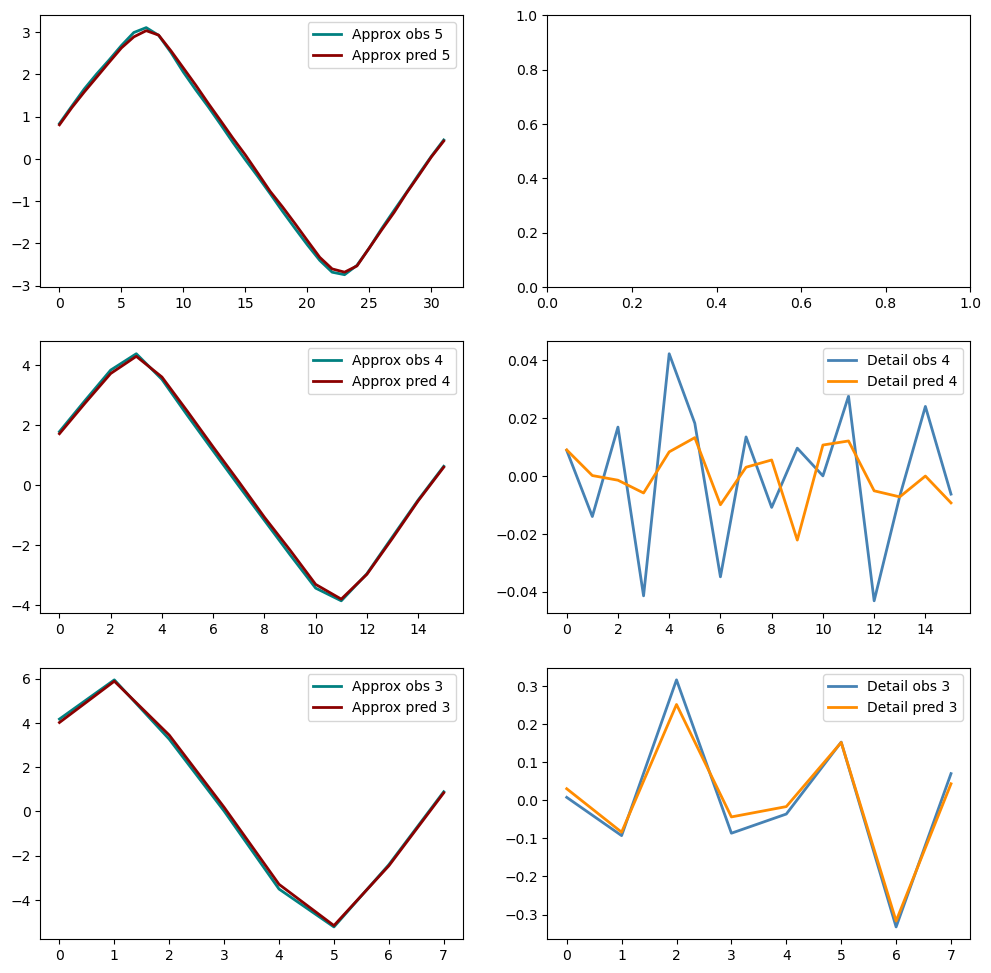}}}
	\subfloat[db$8$]{{\includegraphics[width=0.5\textwidth]{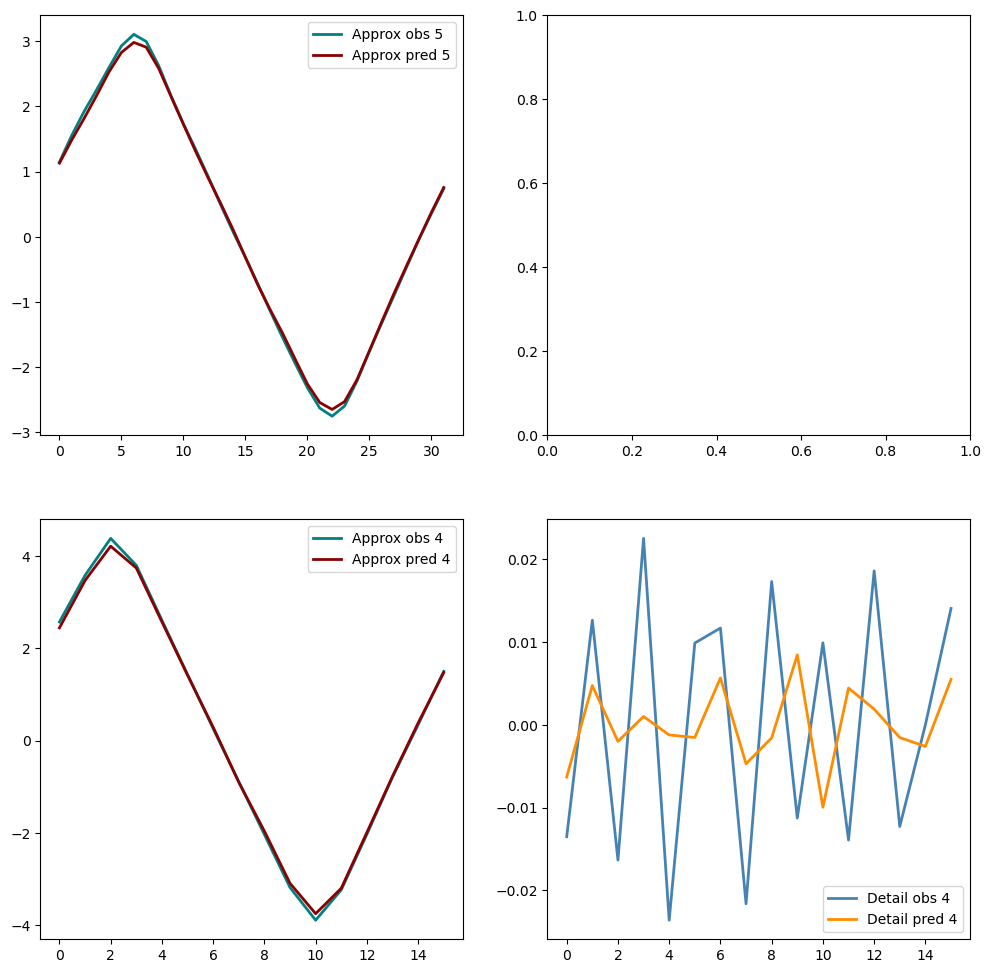}}}	
	\caption{Predicted and observed wavelet decompositions of the first component of the contour depicted in the first column in 
	Figure \ref{fig:spleen_pred}. For db$4$ and db$8$ the detail coefficients on higher resolution levels are too small to be accurately
	predicted, but have little impact on the accuracy of the final approximation.}
	\label{fig:spleen_wav}
\end{figure}

\newpage
\subsubsection{Prostate}
\begin{figure}[htb]
	\centering
	\subfloat[\centering Dice]{{\includegraphics[width=0.5\textwidth]{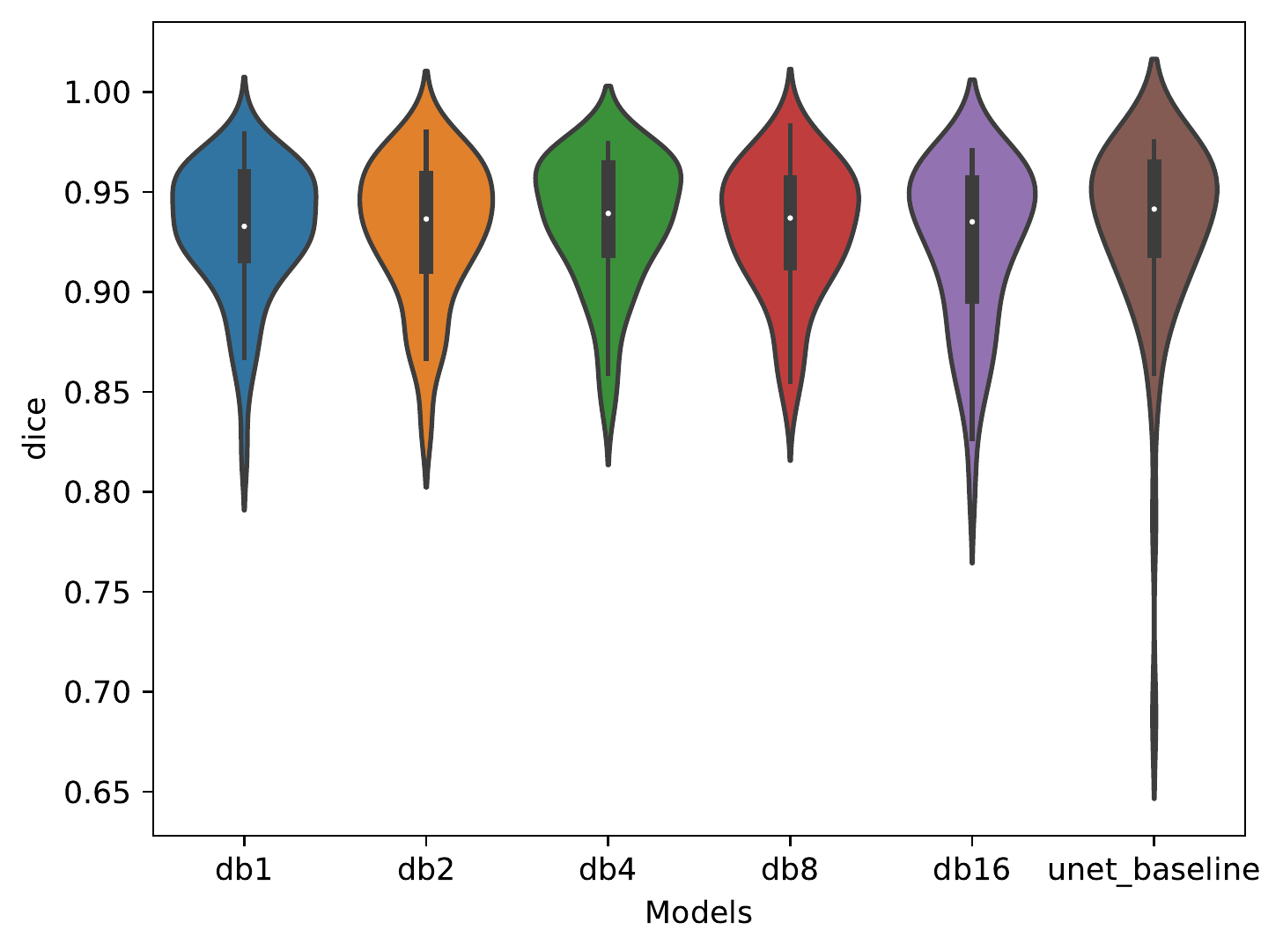}}}
	\subfloat[\centering Hausdorff]{{\includegraphics[width=0.48\textwidth]{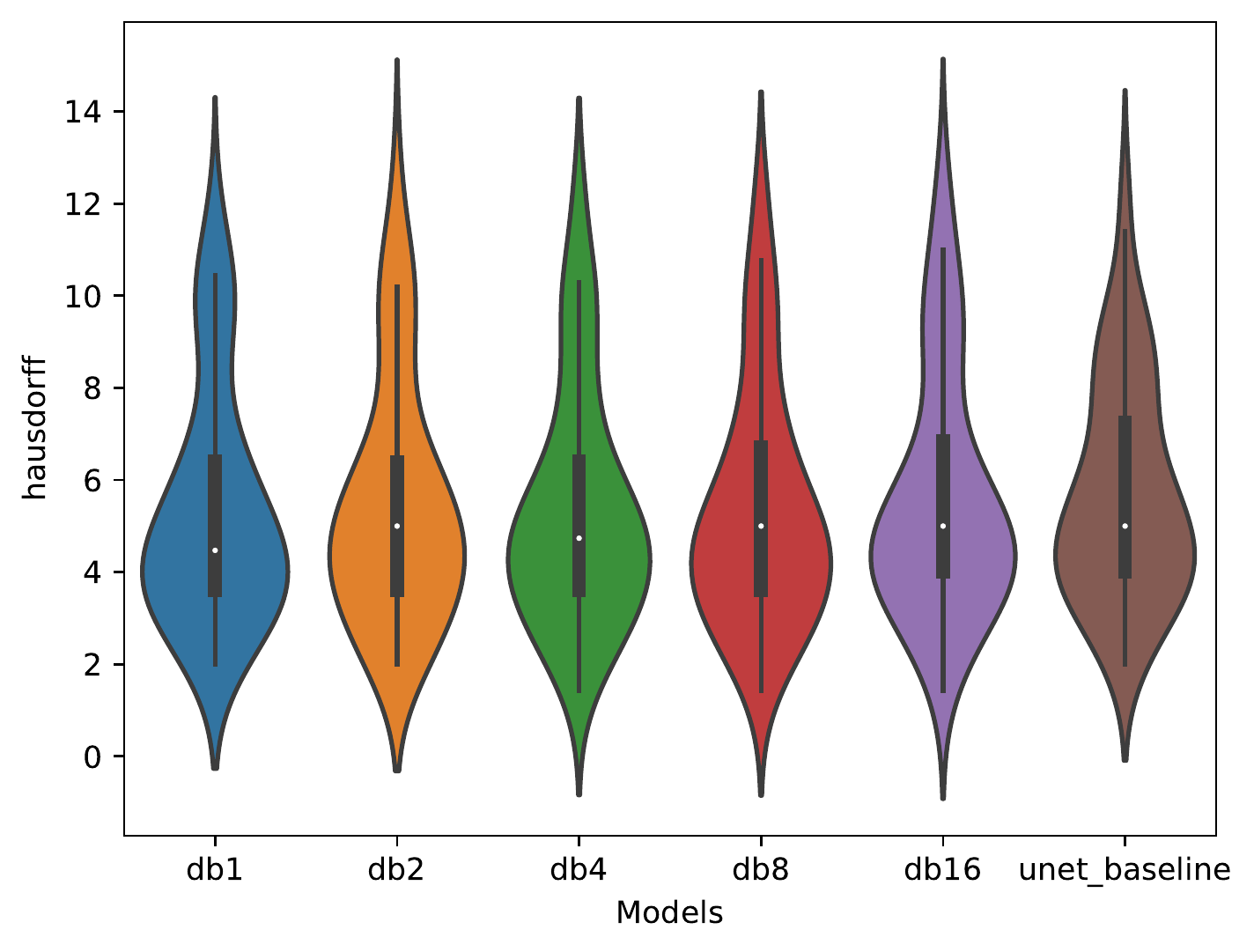}}} \\
	\caption{Boxplots and visualization of approximate densities for the dice scores and Hausdorff distances for the
	prostate.}
	\label{fig:prostate_boxplot}
\end{figure}

\begin{figure}[!htbp]
	\centering
	\subfloat[]{{\includegraphics[width=1\textwidth]{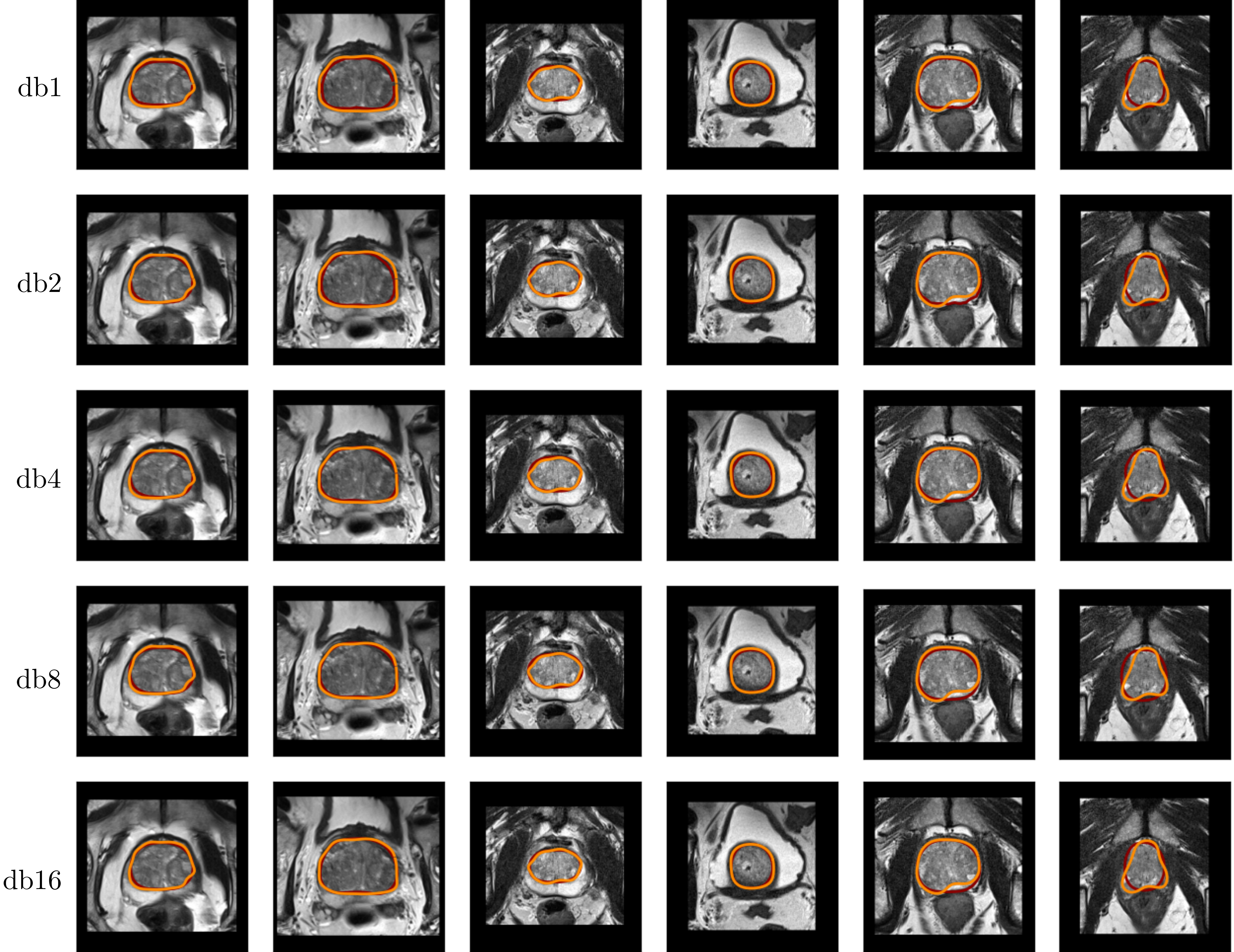}}}
	\caption{Predicted and observed boundaries colored in red and orange, respectively, of the prostate. The
		      last two columns correspond to ``hard'' examples.}
	\label{fig:prostate_pred}
\end{figure}

\begin{figure}[!htbp]
	\centering
	\subfloat[db$1$, fifth column]{{\includegraphics[width=0.5\textwidth]{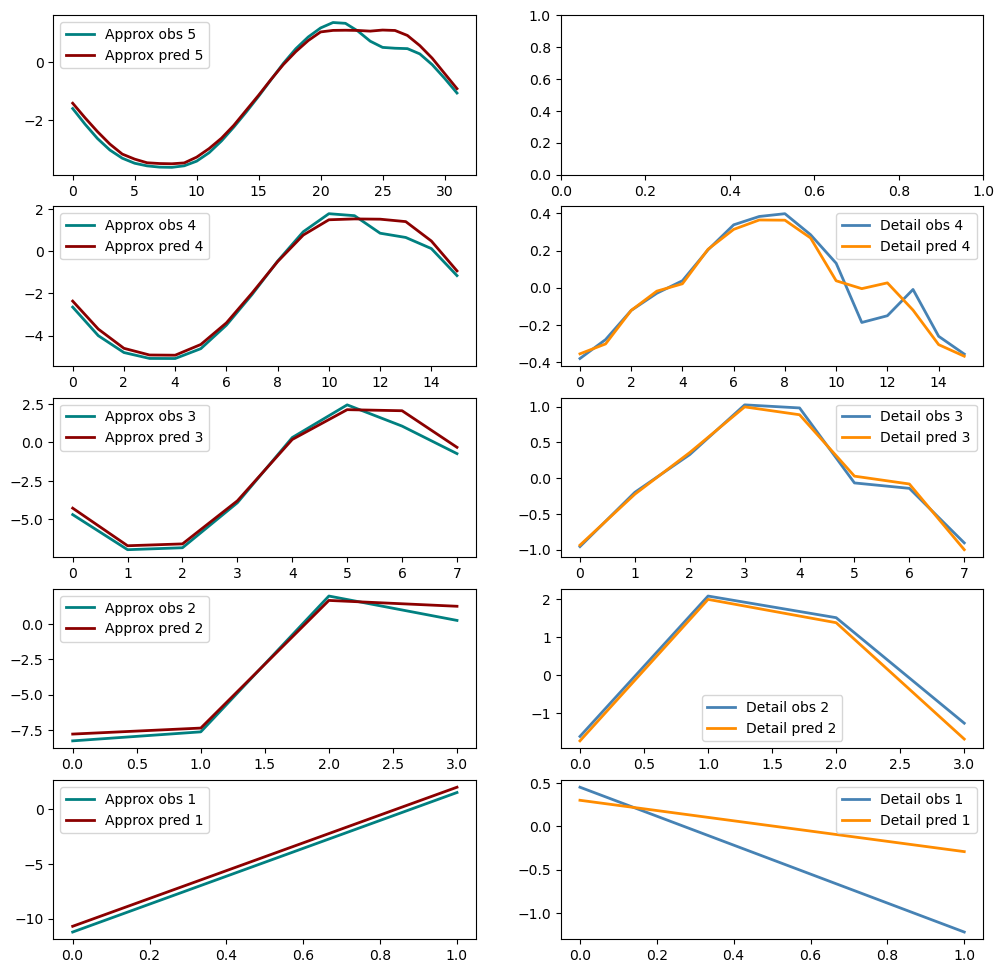}}\label{fig:db1_col5}}
	\subfloat[db$2$, fifth column]{{\includegraphics[width=0.5\textwidth]{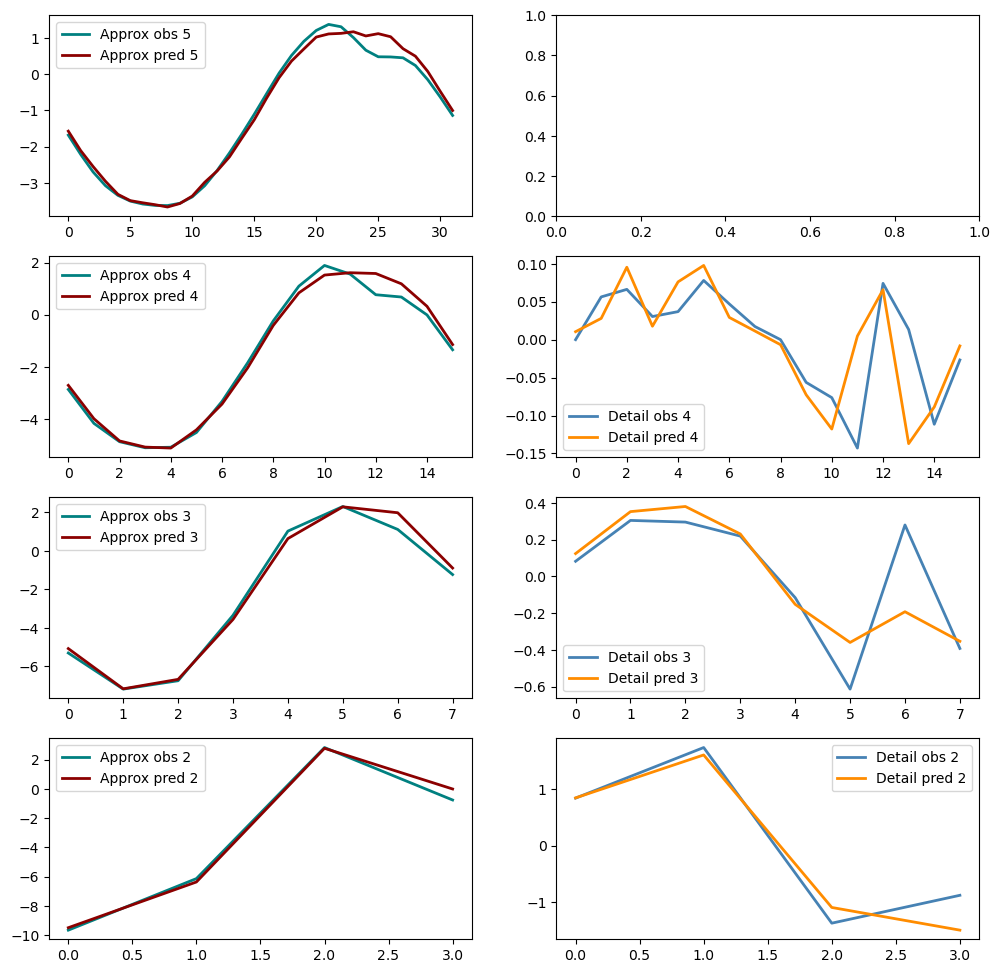}}\label{fig:db2_col5}} \\
	\subfloat[db$1$, sixth column]{{\includegraphics[width=0.5\textwidth]{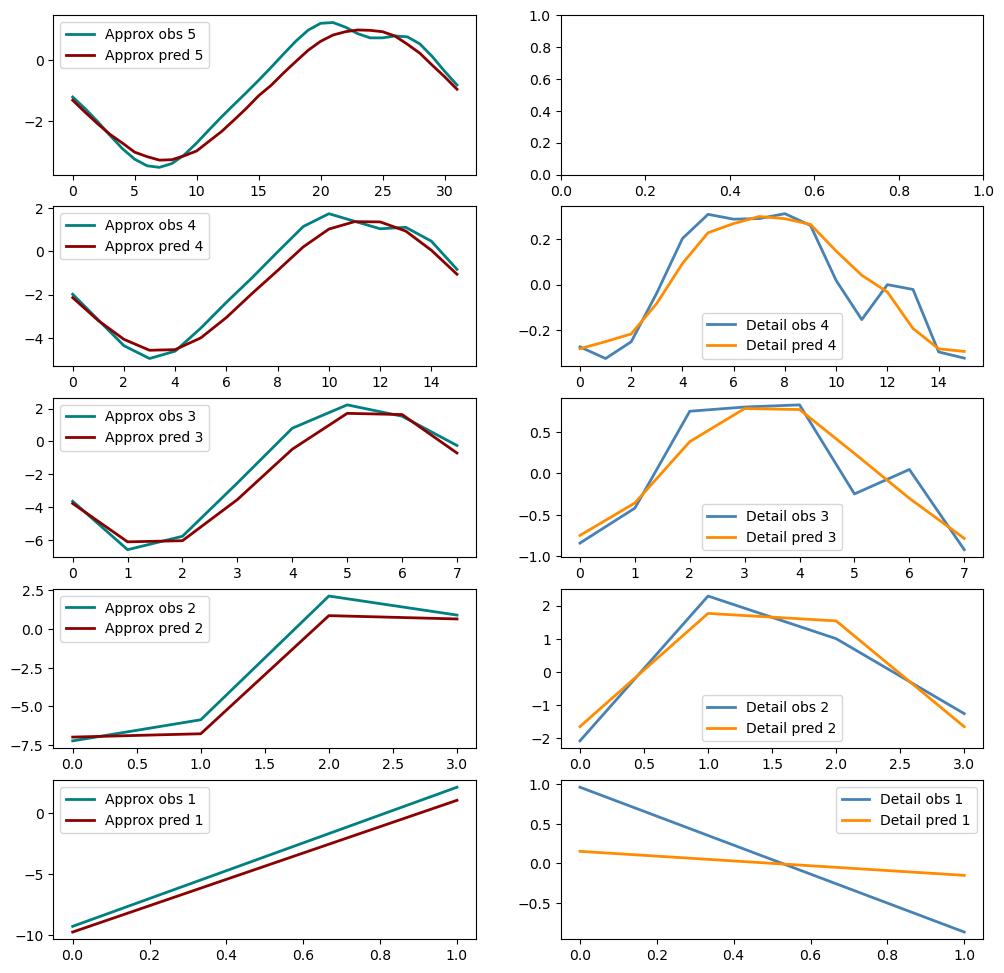}}\label{fig:db1_col6}}
	\subfloat[db$2$, sixth column]{{\includegraphics[width=0.5\textwidth]{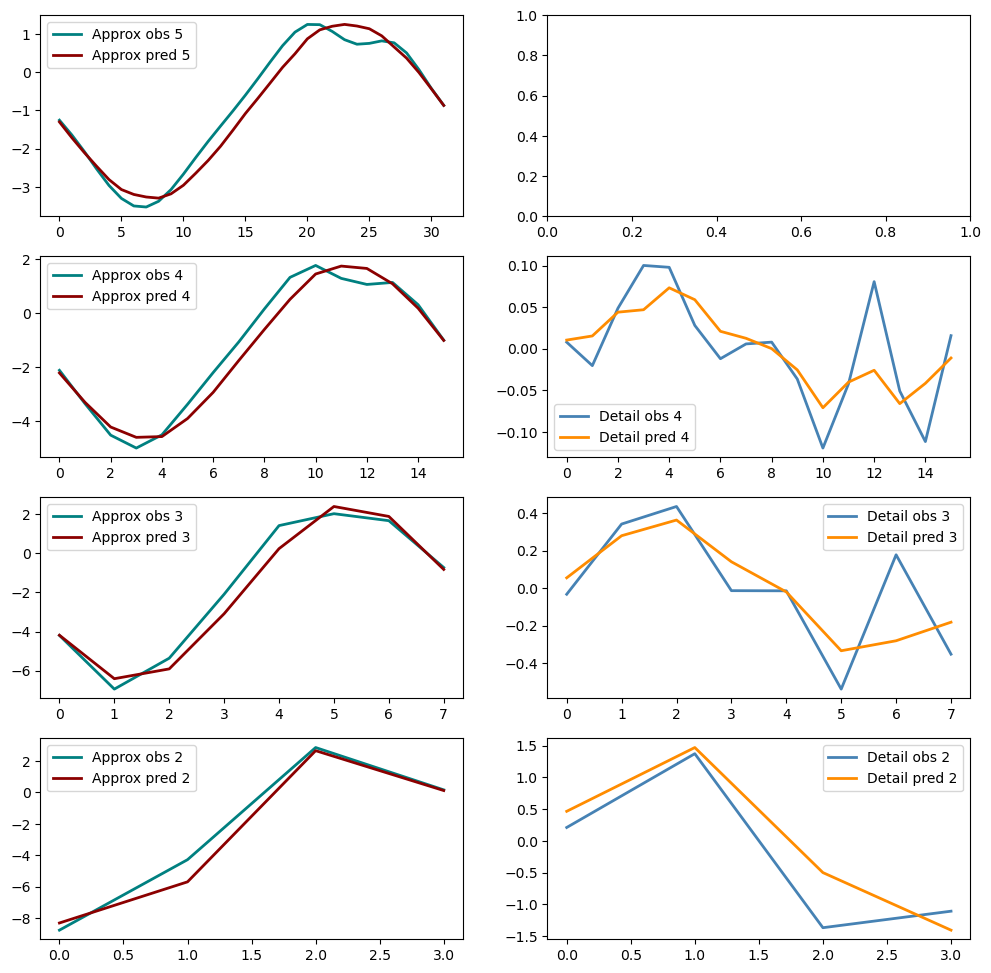}}\label{fig:db2_col6}}	
	\caption{Predicted and observed wavelet decompositions of the first component of the hard examples depicted in
	~\protect \subref{fig:db1_col5}, ~\protect \subref{fig:db2_col5} the fifth column and 
	~\protect \subref{fig:db1_col6}, ~\protect \subref{fig:db2_col6} the sixth column 
	 of Figure \ref{fig:prostate_pred}.}
	\label{fig:prostate_wav}
\end{figure}

\end{document}